\documentclass[letterpaper]{article} % DO NOT CHANGE THIS
\usepackage{aaai2026} 
\usepackage{times}  % DO NOT CHANGE THIS
\usepackage{helvet}  % DO NOT CHANGE THIS
\usepackage{courier}  % DO NOT CHANGE THIS
\usepackage[hyphens]{url}  % DO NOT CHANGE THIS
\usepackage{graphicx} % DO NOT CHANGE THIS
\usepackage{amsmath}
\usepackage{natbib}  % DO NOT CHANGE THIS AND DO NOT ADD ANY OPTIONS TO IT
\usepackage{caption} % DO NOT CHANGE THIS AND DO NOT ADD ANY OPTIONS TO IT
\usepackage{algorithm}
\usepackage{placeins}
\usepackage{booktabs}  % for professional looking tables
\usepackage{tabularx}  % for automatically adjusting table width
\usepackage{multirow}  % for multirow in table
\usepackage[table,xcdraw]{xcolor}
\usepackage[table]{xcolor}
\usepackage{algpseudocode}
\usepackage{makecell}
\usepackage{amsmath}
\usepackage{algpseudocode}
\usepackage{amssymb}
\usepackage{dsfont}
\usepackage[most]{tcolorbox}
\usepackage[table]{xcolor}

\usepackage{newfloat}
\usepackage{listings}
\DeclareCaptionStyle{ruled}{labelfont=normalfont,labelsep=colon,strut=off} % DO NOT CHANGE THIS
\floatstyle{ruled}
\newfloat{listing}{tb}{lst}{}
\floatname{listing}{Listing}
\title{FineXtrol: Controllable Motion Generation via Fine-Grained Text}
\author {
    Keming Shen\textsuperscript{\rm 1,2},
    Bizhu Wu\textsuperscript{\rm 1,2,3},
    Junliang Chen\textsuperscript{\rm 4},
    Xiaoqin Wang\textsuperscript{\rm 1,2},
    Linlin Shen\textsuperscript{\rm 1,2,5}\thanks{Corresponding author.}
}
\affiliations {
    \textsuperscript{\rm 1}School of Computer Science and Software Engineering, Shenzhen University\\
    \textsuperscript{\rm 2}Guangdong Provincial Key Laboratory of Intelligent Information Processing, Shenzhen University\\
    \textsuperscript{\rm 3}School of Computer Science, University of Nottingham Ningbo China, Ningbo, China\\
    \textsuperscript{\rm 4}Department of Electrical and Electronic Engineering, The Hong Kong Polytechnic University\\
    \textsuperscript{\rm 5}Computer Vision Institute, School of Artificial Intelligence, Shenzhen University\\
    2400101005@mails.szu.edu.cn, 
    llshen@szu.edu.cn
}
\usepackage{bibentry}
\begin{document}

\maketitle

% \twocolumn[{%
% \renewcommand\twocolumn[1][]{#1}%
% \maketitle
% \begin{center}
%     \centering
%     \captionsetup{type=figure}
%     \setlength{\abovecaptionskip}{3pt}  % Space above the caption
%     \setlength{\belowcaptionskip}{0pt}  % Space below the caption    
%     \includegraphics[width=1.0\linewidth]{img/0731_fig1_new.pdf}
%     \captionof{figure}{
%         Illustrations of (A) \textbf{Vanilla text-to-motion} methods struggle to control specific body part movements.
%         (B) \textbf{Fine-grained text-to-motion} approaches using LLM-generated detailed descriptions for fine details, but often misalign with ground-truth motions and lack explicit temporal cues.
%         (C) \textbf{Spatially controllable motion generation} methods rely on global 3D coordinate sequences as extra control signals, which are difficult to specify beyond existing datasets and incur high computational costs due to pose conversion.
%         (D) Our \textbf{FineXtrol} introduces accurate and temporally explicit fine-grained textual control signals for specific body parts, enabling user-friendly and efficient controllable motion generation.
%     }
% \label{fig:intro_framework}
%     % \vspace{2.0em}
% \end{center}%
% }]

%%%%%%%%%%%%%%%%%%%%%%%%%%

\begin{abstract}
Recent works have sought to enhance the controllability and precision of text-driven motion generation. 
Some approaches leverage large language models (LLMs) to produce more detailed texts, while others incorporate global 3D coordinate sequences as additional control signals. 
However, the former often introduces misaligned details and lacks explicit temporal cues, and the latter incurs significant computational cost when converting coordinates to standard motion representations.
To address these issues, we propose FineXtrol, a novel control framework for efficient motion generation guided by temporally-aware, precise, user-friendly, and fine-grained textual control signals that describe specific body part movements over time. 
In support of this framework, we design a hierarchical contrastive learning module that encourages the text encoder to produce more discriminative embeddings for our novel control signals, thereby improving motion controllability.
Quantitative results show that FineXtrol achieves strong performance in controllable motion generation, while qualitative analysis demonstrates its flexibility in directing specific body part movements. 
% Code will be released.
\end{abstract}

\section{Introduction}
\label{sec:intro}

\begin{figure*}[!t]
    \centering
    \captionsetup{type=figure}
    \setlength{\abovecaptionskip}{3pt}  % Space above the caption
    \setlength{\belowcaptionskip}{0pt}  % Space below the caption    
    \includegraphics[width=1.0\linewidth]{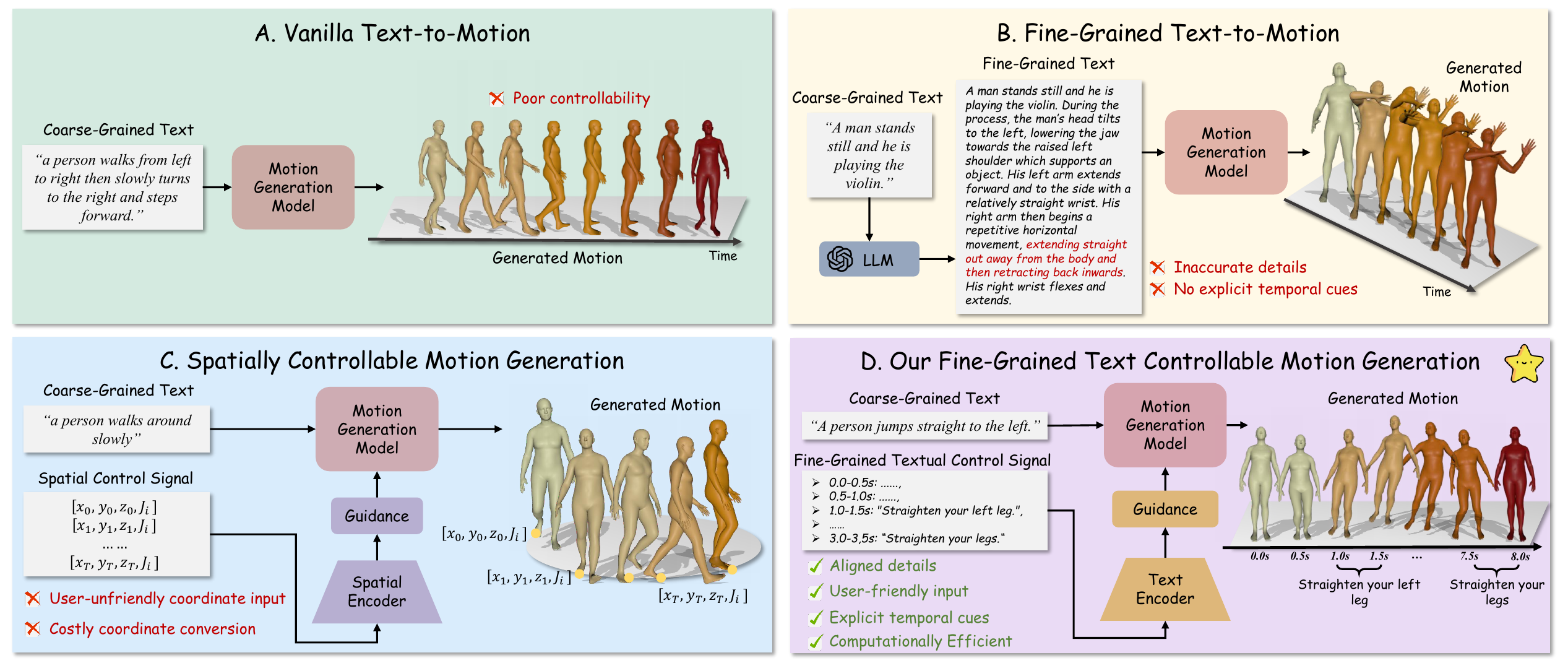}
    \captionof{figure}{
        Illustrations of (A) \textit{Vanilla text-to-motion} methods struggle to control specific body part movements.
        (B) \textit{Fine-grained text-to-motion} approaches using LLMs' expanded descriptions for fine details, but often misalign with ground-truth motions and lack explicit temporal cues.
        (C) \textit{Spatially controllable motion generation} methods rely on global 3D coordinate sequences as extra control signals, which are difficult to be provided beyond existing datasets and incur high computational costs from pose conversion.
        (D) Our \textit{FineXtrol} introduces accurate and temporally explicit fine-grained textual control signals for specific body parts, enabling user-friendly and efficient controllable motion generation.
    }
\label{fig:intro_framework}
    % \vspace{2.0em}
\end{figure*}

% ----------------------------------------------------------------------

Human motion generation~\cite{Action2motion_ACMMM2020,TGcomplete_TOG2020, SDEDiffusion_ICLR2021, MLD_CVPR2023, MotionGPT_NIPS2023} has become an essential task for various applications such as animation and digital humans.
Recent advancements~\cite{Tm2t_ECCV2022,T2M_CVPR2022,MotionDiffuse_PAMI2024,mdm_ICLR2023, MoMask_CVPR2024, sopo_tan2024,foundation_wang2025} have shown impressive capability in synthesizing diverse and realistic human motions from given texts, as illustrated in Fig.~\ref{fig:intro_framework}(A). 
As the quality of generated motions improves, there is a growing demand for generating motions that are not only realistic but also \textbf{precise} and \textbf{controllable}.
Existing works to address these demands fall into two main categories.

On the one hand, some methods~\cite{ActionGPT_ICME2023, CoMo_ECCV2024, wang2025fgt2m++} leverage detailed textual descriptions to generate precise motions, as shown in Fig.~\ref{fig:intro_framework}(B). 
Specifically, large language models (LLMs) are used to expand the coarse-grained textual annotations (\textit{e.g.}, ``A man stands still and he is playing the violin.'') 
into descriptions detailing the movements of individual body parts. 
These expanded fine-grained descriptions are then connected with the original coarse texts to generate motions. 
However, this approach suffers from several limitations. 
First, \textbf{the expanded descriptions are often not strictly aligned with the ground-truth motions}, as noted in~\cite{Wu_FineMotion_ICCV2025}, and may contain inaccurate information due to LLM bias. 
Moreover, the expanded descriptions detail all body part movements of the whole motion sequence, \textbf{lacking explicit temporal cues}, such as \textit{when to raise a hand}, making it difficult to achieve precise control over motion within specific time intervals. 
Besides, this \textbf{simple connection paradigm provides limited flexibility for temporally localized control}, as shown in Tab.~\ref{table:Ablation_FG_MDM}.

On the other hand, methods like~\cite{OmniControl_ICLR2024, InterControl_NIPS2024} introduce spatial control signals to improve controllability, as illustrated in Fig.~\ref{fig:intro_framework}(C).
Inpainting-based methods~\cite{PriorMDM_ICLR2024, mdm_ICLR2023, GMD_CVPR2023} integrated spatial constraints into the generation process, but struggle to control joints beyond the pelvis due to the use of relative pose representations.
Recent methods~\cite{OmniControl_ICLR2024,InterControl_NIPS2024} addressed the above limitations by converting the generated motion to global coordinates, allowing direct comparison with input control signals and using gradient-based error refinement.
However, this coordinate transformation \textbf{incurs significant computational cost}, and specifying realistic global coordinate sequences is challenging and unintuitive for users, \textbf{limiting applicability and scalability}.

To combine the strengths of both categories while addressing their limitations, we propose a novel controllable motion generation framework: FineXtrol—\underline{Fine}-grained te\underline{X}t con\underline{trol}lable motion generation, as displayed in Fig.\ref{fig:intro_framework}(D). 
Instead of relying on spatial coordinate sequences, FineXtrol uses fine-grained textual descriptions of individual body part movements (\textit{e.g.}, ``\textit{Move your left hand to your left thigh.}") as the control signals. 
This eliminates the need for coordinate conversion, improving efficiency, and offers better scalability through the more user-friendly form of control.
Moreover, our fine-grained textual control signals are sourced from FineMotion~\cite{Wu_FineMotion_ICCV2025}, whose descriptions are explicitly aligned with ground-truth motions and encode explicit temporal information.
Besides, rather than directly connecting fine-grained textual control signals with coarse-grained texts as a single input, we incorporate the control signals as residual guidance to modulate the motion features conditioned on coarse-grained texts.
These designs lead to the generation of realistic and coherent motions that adhere closely to the specified fine-grained constraints.

During implementation, we observed that widely used text encoders such as CLIP~\cite{CLIP_ICML2021} and T5~\cite{T5_JMLR2020} struggle to produce discriminative embeddings for our fine-grained textual control signals, as illustrated in Fig.~\ref{fig:text_encoder_comparison}.
We attribute this limitation to their pretraining on datasets that emphasize coarse semantics.
As a result, these encoders often overlook subtle cues essential for capturing detailed motion semantics.
To address this issue, we analyze the characteristics of our fine-grained textual control signals, and introduce a hierarchical contrastive learning module. 
This tailored module enhances the text encoder’s ability to extract discriminative embeddings for our fine-grained textual control signals, thereby enabling more precise control over human motion generation.

Both quantitative results on HumanML3D~\cite{T2M_CVPR2022} and visualizations under various settings demonstrate that our framework, FineXtrol, delivers strong controllable motion generation performance and notably surpasses the previous state-of-the-art in controlling multiple body parts within designated temporal intervals. 
Moreover, compared with prior diffusion-based controllable motion generation methods, FineXtrol uses fewer trainable parameters and requires less inference time, highlighting the efficiency of our framework.
% Experimental results on HumanML3D~\cite{T2M_CVPR2022} under various settings demonstrate that our proposed framework, FineXtrol, achieves promising performance in controllable motion generation, notably surpassing the previous state-of-the-art in controlling combinations of body parts.
% Moreover, FineXtrol is efficient and offers significant convenience for users. 
% Visualizations further illustrate its capability to control specific body parts, adjust their movements within specific intervals, and simultaneously manipulate multiple body parts via our fine-grained textual control signals.
To summarize, our contributions are as follows:
\textbf{First}, to the best of our knowledge, we propose FineXtrol, a novel and user-friendly controllable motion generation framework that enables control over body part movements within specific temporal intervals.
\textbf{Secondly}, we design a hierarchical contrastive learning module to enhance the text encoder’s ability to capture discriminative embeddings for our fine-grained textual control signals, thereby further improving FineXtrol’s capacity to generate motions that accurately follow the specified instructions.
\textbf{Thirdly}, extensive experiments demonstrate that our FineXtrol excels in precise multi-body-part motion control, and user accessibility, highlighting its potential for real-world applications.

%-------------------------------------------------------------------

\begin{figure*}[!t]
\begin{center}
\includegraphics[width=0.9\linewidth]{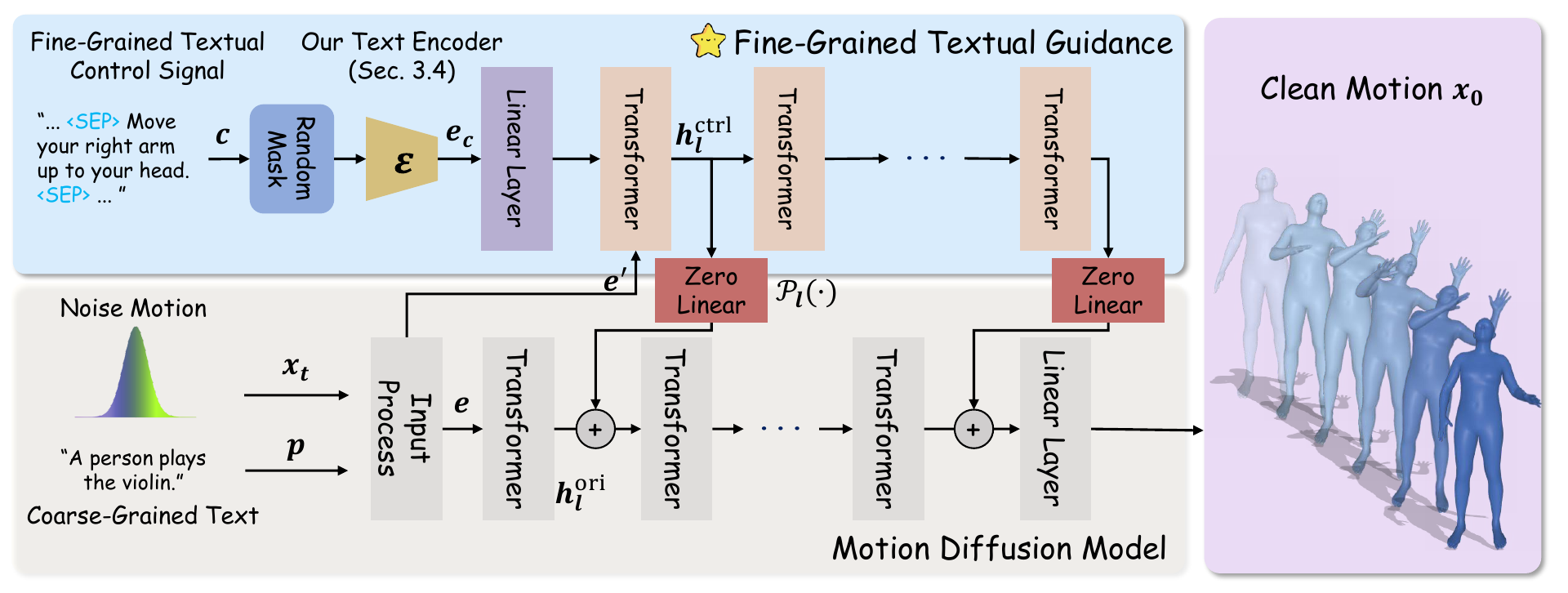}
\end{center}
\setlength{\abovecaptionskip}{-0em}  % Space above the caption
\caption{
    Overview of FineXtrol.
    Our framework takes the coarse-grained text $\boldsymbol{p}$, the fine-grained textual control signal $\boldsymbol{c}$, and a noise motion sequence $\boldsymbol{x_t}$ as input, and predicts the clean motion sequence $\boldsymbol{x_0}$. 
    The lower branch resumes from MDM to maintain stable motion generation capabilities from $\boldsymbol{p}$.
    The upper branch is a trainable copy of MDM, modulated by $\boldsymbol{c}$ through conditional feature adaptation. 
    The zero-initialized linear layers connect between branches. 
    The framework ensures that the generated motion adheres not only to the coarse-grained text but also to the control signal.}
\label{fig:model_overview}
\end{figure*}

\section{Related Work}
\label{sec:related_work}
\subsubsection{Text-driven Human Motion Generation.}
Text-to-motion generation plays a crucial role in various applications. Existing approaches can be broadly categorized into two major generative paradigms~\cite{OmniControl_ICLR2024}:
auto-regressive models~\cite{T2M_CVPR2022, MoMask_CVPR2024, DeepPhase_TOG2022, TRACE_CVPR2023, CoMo_ECCV2024, PADL_SIGGRAPH2022} and diffusion-based models~\cite{mdm_ICLR2023,Physdiff_ICCV2023,PriorMDM_ICLR2024,GMD_CVPR2023,MLD_CVPR2023}.
Auto-regressive approaches generate motion frame-by-frame, while diffusion-based ones progressively denoise an entire motion sequence over multiple iterations. 
Most of them rely on coarse-grained texts, which often fail to capture the subtle details of human actions.
Recent studies~\cite{ActionGPT_ICME2023,CoMo_ECCV2024,wang2025fgt2m++} have incorporated fine-grained textual descriptions to enhance precise motion generation.
However, their LLM-generated detailed texts are often not strictly aligned with ground-truth motions and lack explicit temporal cues, leading to weak correspondence between text and motion segments.
This limits control over individual body parts and time intervals.
In contrast, our FineXtrol enables precise control over both.

%-------------------------------------------------------------------

% \vspace{0.5em}
\subsubsection{Controllable Motion Diffusion Models.}
Controllability in diffusion models has been widely studied, especially in image synthesis~\cite{LDM_CVPR2022,Classifier_free,ControlNet_ICCV2023,wu2024light}, and is now also gaining attention in motion generation area.
Inpainting-based methods like PriorMDM~\cite{PriorMDM_ICLR2024} condition on observed motion segments to predict missing ones while maintaining global coherence.
ControlNet-based methods enhance controllability by injecting conditional signals via a trainable copy of a pretrained motion generation model. 
For example, OmniControl~\cite{OmniControl_ICLR2024} and InterControl~\cite{InterControl_NIPS2024} use global joint coordinates as control signals, enabling controllability with minimal fine-tuning. 
However, they require users to provide realistic coordinate sequences, which is impractical, and incur expensive conversions to relative motion representations.
% limiting scalability and generalization.
Similarly, our method adopts ControlNet paradigm, but employs a more efficient and user-friendly control signal, \textit{i.e.}, fine-grained textual descriptions of body part movements across temporal intervals.

%-------------------------------------------------------------------------
% \vspace{0.5em}
\subsubsection{Contrastive Learning for Text Embedding.}
Learning discriminative sentence embeddings is fundamental to natural language processing. 
Early methods relied on distributional hypotheses to predict surrounding sentences~\cite{Skip_Thought_NIPS2015, QuickThoughts_ICLR2018, ACTOR_ICCV2021} or extended Word2Vec with N-Gram representations~\cite{Sent2Vec_NAACL2018}.
Recent methods have adopted contrastive learning to align augmented sentence pairs~\cite{BERT_NAACL2019, DECLUTR_IJCNLP2021, CT_ICLR2021, CSGN_ICCV2019, SimCSE_EMNLP2021}.
Contrastive learning has also been effective in multi-modal settings~\cite{BEiT_3_Microsoft2022, DALL_E_IMCL2021, CLIP_ICML2021, DisFaceRep_MM2025, Flip_MM2024}, aligning image and text embeddings from large-scale paired data.
However, most pretrained text encoders focus on coarse semantics and struggle with fine-grained distinctions.
To address this, we analyze the properties of our fine-grained textual control signals, and design a contrastive module to extract discriminative embeddings, enabling more precise fine-grained text controllable motion generation.

\section{Method}
\label{sec:method}

Fig.~\ref{fig:model_overview} displays the overall pipeline of our proposed framework, which leverages fine-grained textual descriptions as control signals to guide motion generation.
The notations and the definition of this novel research problem are clarified in Sec.~\ref{sec:Formulation}.
We then briefly introduce required preliminaries in Sec.~\ref{sec:Background}.
Next, we detail two key modules in our framework: 
(1) generating motion sequences guided by our fine-grained textual control signals, which detail specific body part movements in specific temporal intervals (see Sec.~\ref{sec:Motion ControlNet}),
and (2) representing our fine-grained textual control signals in a robust and discriminant manner (see Sec.~\ref{sec:CL}).

%-------------------------------------------------------------------------

\subsection{Problem Formulation}
\label{sec:Formulation}
Given a coarse-grained text $\boldsymbol{p}$, such as ``\textit{A man kicks something with his left leg}'', and a fine-grained textual control signal $\boldsymbol{c}$, such as ``\textit{Move your left leg to the right in 1.0-1.5s}'',
our framework $\mathcal{F}$ should generate a realistic motion sequence $\boldsymbol{x_0} \in \mathbb{R}^{T \times D}$, where $T$ is the temporal length and $\mathcal{D}$ is the dimension of human pose representations. 
Mathematically, we formulate this fine-grained text controllable motion generation task as:
\begin{equation}
\boldsymbol{x_0}=\mathcal{F}\left(\boldsymbol{p}, \boldsymbol{c}; \Theta \right),
\end{equation}
where $\Theta$ is the parameters of $\mathcal{F}$.

%------------------------------------------------------------------------

\subsection{Preliminaries}
\label{sec:Background}

\noindent\textbf{Motion Diffusion Model (MDM)} ~\cite{mdm_ICLR2023} adapted powerful diffusion models to human motion generation by framing the task as a gradual denoising process. 
Given a ground-truth motion $\boldsymbol{\hat{x}_0}$, noise is progressively added to produce a noisy version $\boldsymbol{x_t}$, where $t$ denotes the number of noise addition steps.
Then, conditioned on a time step $t$ and a coarse-grained text $\boldsymbol{p}$, MDM employed a transformer-based network $\epsilon_\theta$ with parameters $\theta$ to reverse the process and directly predict the clean motion $\boldsymbol{x_0}$. 
The model is trained with the objective function:
\begin{equation}
\mathcal{L}_\theta =\left\| \epsilon_\theta(\boldsymbol{x}_{t}, t, \boldsymbol{p}; \theta)-\boldsymbol{\hat{x}_0}\right\|_2^2.
\end{equation}

\begin{figure}[H]
\begin{center}
\includegraphics[width=1.0\linewidth]{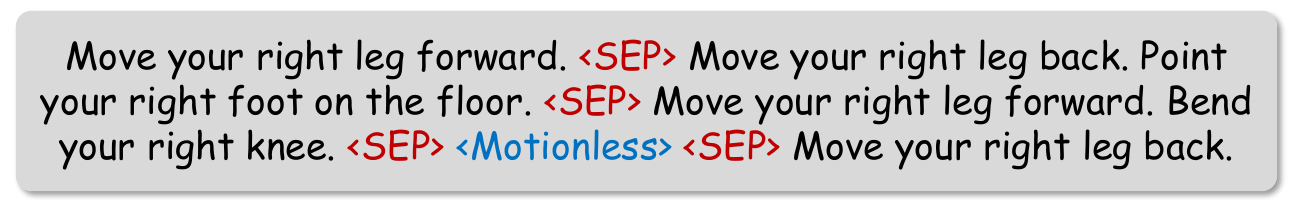}
\end{center}
\setlength{\abovecaptionskip}{-0em}  % Space above the caption
\caption{
    A fine-grained textual control signal example.
}
\label{fig:dt_example}
\end{figure}

% \vspace{0.5em}
\noindent\textbf{Fine-grained Textual Control Signal} $\boldsymbol{c}$ specifies the desired movements of particular body parts within defined temporal intervals. 
In this paper, we divide the human body into six main parts: \textit{head}, \textit{body}, \textit{left arm}, \textit{right arm}, \textit{left leg}, and \textit{right leg}.
See the Appendix for details.
An example of a fine-grained control signal for the \textit{right leg} is shown in Fig.~\ref{fig:dt_example}. 
Here, \textless SEP\textgreater\,separates the descriptions across different temporal intervals. \textless Motionless\textgreater\,indicates that no movement of the specified body part occurs in that interval.

%%%%%%%%%%%%%%%%%%%%%%%%%%%%%%%%%%%%%%%%%

\subsection{Motion Generation with Fine-grained Textual Control Signal}
\label{sec:Motion ControlNet}

To generate human motions that follow coarse-grained texts and the fine-grained textual control signals, we follow the ControlNet~\cite{ControlNet_ICCV2023} paradigm to propose a dual-branch framework, as illustrated in Fig.~\ref{fig:model_overview}. 
The \textbf{lower branch} reuses the original MDM~\cite{mdm_ICLR2023} transformer encoder with resumed pretrained weights to ensure stable generation from coarse-grained texts $\boldsymbol{p}$. 
Specifically, the ``\textit{Input Process}'' block includes a CLIP~\cite{CLIP_ICML2021} text encoder to extract the textual embedding $\boldsymbol{e}_p$, and linear layers to encode the noisy motion sequence $\boldsymbol{x_t}$ into a motion embedding $\boldsymbol{e}_{x_t}$. 
These two embeddings are then concatenated as
\begin{equation}
\boldsymbol{e} = [\boldsymbol{e}_p ; \boldsymbol{e}_{x_t}],
\end{equation}
and fed into the transformer blocks.
The output of the $l$-th transformer block is defined as
\begin{equation}
\boldsymbol{h}_l^{\text{ori}} = \text{TransformerBlock}_l(\boldsymbol{h}_{l-1}^{\text{ori}}), \quad \text{with } \boldsymbol{h}_0^{\text{ori}} = \boldsymbol{e}.
\end{equation}

Meanwhile, the \textbf{upper branch} is a trainable copy of MDM, modulated by the fine-grained textual control signal $\boldsymbol{c}$ through conditional feature adaptation.
First, we construct the concatenated textual and motion embedding $\boldsymbol{e}'$ in the same way as the lower branch.
To encode the control signal $\boldsymbol{c}$, we apply random masking on it by replacing descriptions within random temporal intervals with the special token \textless Mask\textgreater, and extract its embedding $\boldsymbol{e}_c$ using our well-designed text encoder (see Sec.~\ref{sec:CL}).
We then align $\boldsymbol{e}_c$ to the transformer’s embedding space via a linear layer and add it to $\boldsymbol{e}'$, effectively modulating $\boldsymbol{e}'$ before it is fed into the transformer.
The output of the $l$-th transformer block in the upper branch is computed as:
\begin{equation}
\boldsymbol{h}_l^{\text{ctrl}} = \text{TransformerBlock}_l(\boldsymbol{h}_{l-1}^{\text{ctrl}}),
\end{equation}
\begin{equation}
\text{with } \boldsymbol{h}_0^{\text{ctrl}} = \boldsymbol{e}' + \text{Linear}(\boldsymbol{e}_c).
\end{equation}

The upper branch is connected to the lower branch through linear layers $\mathcal{P}_l(\cdot)$ whose weights and biases are initialized to zero. 
Formally, the interaction between branches at the $l$-th layer is defined as:
\begin{equation}
\boldsymbol{h}_l^{\text{out}} = \boldsymbol{h}_l^{\text{ori}} + \mathcal{P}_l(\boldsymbol{h}_l^{\text{ctrl}}).
\end{equation}
% where $\mathcal{P}_l(\cdot)$ denotes the zero-initialized linear layers.
This zero-initialization prevents the injection of random noise into the lower branch during the early stages of training. 
As training progresses, the upper branch gradually learns to interpret the fine-grained textual control signals, capturing both spatial and temporal aspects, and refines the motion by injecting meaningful corrections into the corresponding layers of the motion diffusion model, thereby enhancing motion quality.

%-------------------------------------------------------------------------

\subsection{Hierarchical Contrastive Learning for Fine-Grained Textual Control Signal}
\label{sec:CL}

Commonly used text encoders, such as CLIP~\cite{CLIP_ICML2021} and T5~\cite{T5_JMLR2020}, exhibit limited capability in extracting discriminative embeddings for fine-grained textual descriptions, especially our fine-grained textual control signals. 
This limitation hinders the performance of controllable motion generation. 
To solve this, we analyze the structure of our control signal and identify that it inherently contains three levels of information. 
As illustrated in Fig.~\ref{fig:dt_example}:
% \vspace{0.5em}
\begin{itemize}
    \item The whole example describes a specific body part's movements across all temporal intervals. Such information is referred to as \textbf{sequence-level}.

    \item Descriptions of a single interval, \textit{i.e.}, between two\,\textless SEP\textgreater\,tokens, are considered as \textbf{snippet-level}.

    \item Each individual sentence within an interval is defined as \textbf{sentence-level}.
\end{itemize}

Building on this insight, we propose a hierarchical contrastive learning module with level-specific data augmentations to enhance the control signal embeddings. 
Specifically, we adopt T5 as the base text encoder due to its capacity to encode long text sequences, and progressively train it through contrastive learning at the sentence, snippet, and sequence levels, each initialized from the weights learned at the previous level. 
Next, we describe data augmentations for each level and the training objective in detail.

% \vspace{0.5em}
\subsubsection{Sentence-Level.} 
\label{sec:sentence-level}
At this level, the goal is to enable the text encoder to distinguish between different body part movement sentences.
Specifically, we build a sentence-level corpus $\mathcal{D}_\text{sen}^{\text{ori}}$ with non-repetitive body part movement sentences from FineMotion~\cite{Wu_FineMotion_ICCV2025}, which is the source of our fine-grained textual control signals. 
Then, we utilize DeepSeek-V2~\cite{deepseekv2} to rewrite each sentence, forming an augmented corpus $\mathcal{D}_\text{sen}^{\text{aug}}$ (see Appendix for prompts).
Each sentence in $\mathcal{D}_\text{sen}^{\text{ori}}$ and its rewritten counterpart in $\mathcal{D}_\text{sen}^{\text{aug}}$ form a positive pair, while all other sentences in the corpora serve as their negative samples.

% \vspace{0.5em}
\subsubsection{Snippet-Level.}
This level aims to make the text encoder robust to the order of sentences within a single time interval.
Similarly, we collect snippet-level descriptions from FineMotion to build a snippet-level corpus $\mathcal{D}_\text{sni}^{\text{ori}}$.
For augmentation, we randomly replace some of the sentences with their counterparts from $\mathcal{D}_\text{sen}^{\text{ori}}$ and shuffle their order (see Appendix, Algorithm~\ref{alg:snippet_aug}). 
Each snippet-level description is augmented twice to form a positive pair; augmentations from other snippet-level ones serve as their negative samples.

% \vspace{0.5em}
\subsubsection{Sequence-Level.} 
The text encoder in this level aims to enhance its temporal awareness of different temporal intervals' body part movement descriptions, \textit{i.e.}, our control signals.
To construct augmented data at this level, we randomly apply snippet-level augmentations to the individual intervals within a control signal, but preserving the temporal order of intervals (see Appendix, Algorithm~\ref{alg:seq_aug}).
Each sequence-level description is augmented twice to form a positive pair, while augmented versions of other sequence-level descriptions are treated as their negative samples.

% \vspace{0.5em}
\subsubsection{Contrastive Learning.} 

For all three levels, we train the text encoder with the InfoNCE~\cite{InfoNCE_2018} loss, which pulls positive pairs closer and pushes negatives apart.
In each minibatch, we sample $N$ original texts and construct positive pairs $\boldsymbol{c}^\text{aug}_i$ and $\boldsymbol{c}^\text{aug}_j$ for each.
Their embeddings can be represented as:
\begin{equation}
\label{eqn:text_encoder}
h_i=\text{Avg}(\mathcal{E}\left(\boldsymbol{c}_i^{\mathrm{aug}}\right)), 
\quad 
h_j=\text{Avg}(\mathcal{E}\left(\boldsymbol{c}_j^{\mathrm{aug}}\right)),
\end{equation}
where $\mathcal{E}(\cdot)$ is the text encoder, and $\text{Avg}(\cdot)$ denotes the average pooling operation that aggregates the encoder outputs along the sequence length to produce a single embedding.
A MLP projection head $g(\cdot)$, randomly initialized for each level, then maps the embeddings into contrastive space:
\begin{equation}
    z_i=g(h_i), 
    \quad
    z_j=g(h_j).
\end{equation}
Thus, each minibatch yields $2N$ contrastive embeddings in total.
% During training, we randomly sample a minibatch of $N$ original text, yielding $2N$ augmented samples. 
Each embedding is contrasted against the other $2(N-1)$ negative ones. 
The contrastive loss for each embedding, such as $\boldsymbol{c}^\text{aug}_i$, is:
\begin{equation}
\label{eqn:infoNCE}
\mathcal{L}_i=-\log \frac{\exp \left(\operatorname{sim}\left(\boldsymbol{z}_i, \boldsymbol{z}_j\right) / \tau\right)}{\sum_{k=1}^{2 N} \mathds{1}_{[k \neq i]} \exp \left(\operatorname{sim}\left(\boldsymbol{z}_i, \boldsymbol{z}_k\right) / \tau\right)},
\end{equation}
where $\operatorname{sim}(\cdot, \cdot)$ denotes cosine similarity, $\mathds{1}(\cdot)$ is an indicator function, and $\tau$ is the temperature parameter.

\begin{table*}[!t]
\centering
\footnotesize
\resizebox{1.0\textwidth}{!}
{
\begin{tabular}{c|c|cc|c|cccc}

\toprule

\textbf{No.} & \textbf{Method} & \makecell{\textbf{Control} \\ \textbf{signal}} & \makecell{\textbf{User-Friendly }\\ \textbf{control}} & \textbf{Body Part} & \textbf{FID} $\downarrow$ & \makecell{\textbf{R-precision} $\uparrow$ \\ \textbf{(Top-3)}} & \textbf{Diversity} $\rightarrow$ & \textbf{MM-Dist}  $\downarrow$ \\
\midrule

(1) & Real & - & - & - & 0.002  & 0.796 & 9.503 & 2.965 \\ 

\midrule

(2) & MDM~\cite{mdm_ICLR2023} $_{\text{ICLR'23}}$& - & - & - & 0.544  & 0.611 & 9.559 & 5.432 \\ 

\midrule

\multirow{5}{*}{(3)} & MDM~\cite{mdm_ICLR2023} $_{\text{ICLR'23}}$  & & $ \times $ & \multirow{4}{*}{Pelvis} & 0.698  & 0.602 & 9.197 & 5.430 \\
& PriorMDM~\cite{PriorMDM_ICLR2024} $_{\text{ICLR'24}}$  &  & $ \times $ &  \multirow{4.5}{*}{Only}& 0.475  & 0.583 & 9.156 & 5.424  \\
& GMD~\cite{GMD_CVPR2023} $_{\text{CVPR'23}}$  & Coordinate & $ \times $ &  & 0.576  & 0.665 & 9.206 & 5.430 \\
& OmniControl~\cite{OmniControl_ICLR2024} $_{\text{ICLR'24}}$  &  & $ \times $ &  & 0.218  & 0.687  & 9.422 & 4.991 \\
& InterControl~\cite{InterControl_NIPS2024} $_{\text{NeurIPS
'24}}$  & & $ \times $ &  & 0.159  & 0.671  & 9.482 & 5.026 \\

\midrule\midrule

\multirow{2}{*}{(4)} & InterControl~\cite{InterControl_NIPS2024} $^\dag$ $_{\text{NeurIPS '24}}$ & \multirow{2}{*}{Coordinate} & $ \times $ &  \multirow{2}{*}{Average} & \textbf{0.209}  & \textbf{0.684} & \textbf{9.301} & 5.164 \\
 & OmniControl~\cite{OmniControl_ICLR2024} $^\dag$ $_{\text{ICLR'24}}$ &  & $ \times $ &  & 0.255  & 0.680 & 9.735 & \textbf{5.054} \\

\midrule
 
\multirow{2}{*}{(5)} & CoMo ~\cite{CoMo_ECCV2024} $^\dag$ $_{\text{ECCV'24}}$ & \multirow{2}{*}{Text} & \checkmark & \multirow{2}{*}{Average} & 0.347  & 0.625 & 9.568 & 5.588 \\

 & \textbf{FineXtrol (Ours)} & & \checkmark & & \textbf{0.245} & \textbf{0.685} & \textbf{9.492} & \textbf{5.087} \\

\midrule\midrule

\multirow{3}{*}{(6)} & OmniControl~\cite{OmniControl_ICLR2024} $^{\dag}_{\text{ICLR'24}}$& Coordinate  &$ \times $ &  & 0.624  & 0.601  & 9.334 & 5.252 \\

& CoMo ~\cite{CoMo_ECCV2024} & Text & \checkmark & & 0.606 & 0.611 & 9.662 & 5.638 \\

&  \textbf{FineXtrol (Ours)} & Text & \checkmark & \multirow{-3}{*}{Cross} & \textbf{0.351} & \textbf{0.676} & \textbf{9.658} & \textbf{5.146} \\

\bottomrule

\end{tabular}
} %Line 652

\caption{
Quantitative results on the HumanML3D test set.
Methods in block (3) are trained with pelvis-only control, 
while those in blocks (4)-(6) are trained with control over all body parts.
\textit{Body Part (Average)} reports the average performance across all body parts under four levels of control signal density: 25\%, 50\%, 75\%, and 100\%.
\textit{Body Part (Cross)} reports the average performance over the combination of multiple body parts.
$\dag$ refers to our re-evaluation.
$\rightarrow$ means closer to real data is better. 
Results show that our FineXtrol is efficient, user-friendly, and competitive with the existing controllable generation methods. 
We bold the best results in each block with body part: \textit{Average} and \textit{Cross}.
}
\label{table:main}
\end{table*}

\section{Experiment}
\label{Sec:exp}

\subsection{Experimental Setup}
\subsubsection{Datasets.} 
We conduct experiments on the widely used HumanML3D~\cite{T2M_CVPR2022} dataset, using its texts as the coarse-grained texts $\boldsymbol{p}$.
For fine-grained textual control signals $\boldsymbol{c}$, we use annotations from FineMotion~\cite{Wu_FineMotion_ICCV2025}, which provide detailed body part movement descriptions over short temporal intervals for HumanML3D motions.
See the Appendix for details.

% \vspace{0.5em}
\subsubsection{Implementation Details.} 
We first progressively train the text encoder across three levels in the hierarchical contrastive learning module, then freeze it to extract control signal embeddings for training our FineXtrol framework.
All experiments are conducted on a single A100 40G GPU.
We follow the training hyperparameters from~\cite{OmniControl_ICLR2024}.
% , except for the total training iterations, which are set to 400K (approximately 40 hours).
See the Appendix for detailed hyperparameters.

% \vspace{0.5em}
\subsubsection{Evaluation Metrics.} 
The evaluation metrics are categorized into two groups:
(1) Generated Motion Quality: 
Following~\cite{OmniControl_ICLR2024}, we assess the realism and diversity of generated motions using Frechet Inception Distance (FID), Multi-modal Distance (MM-Dist), R-Precision (Top-1/2/3 motion-to-text retrieval accuracy), and Diversity. 
For metric definitions, see~\cite{T2M_CVPR2022}.
(2) Textual Representation Quality: 
To evaluate the discriminability of embeddings for our fine-grained textual control signals, we adopt metrics from~\cite{BERT_NAACL2019, SimCSE_EMNLP2021}, including cosine similarity, alignment, and uniformity. See their papers for details.

% ----------------------------------------------------------------------

% 实验结果
\begin{figure}[!b]
\begin{center}
\includegraphics[width=1.0\linewidth]{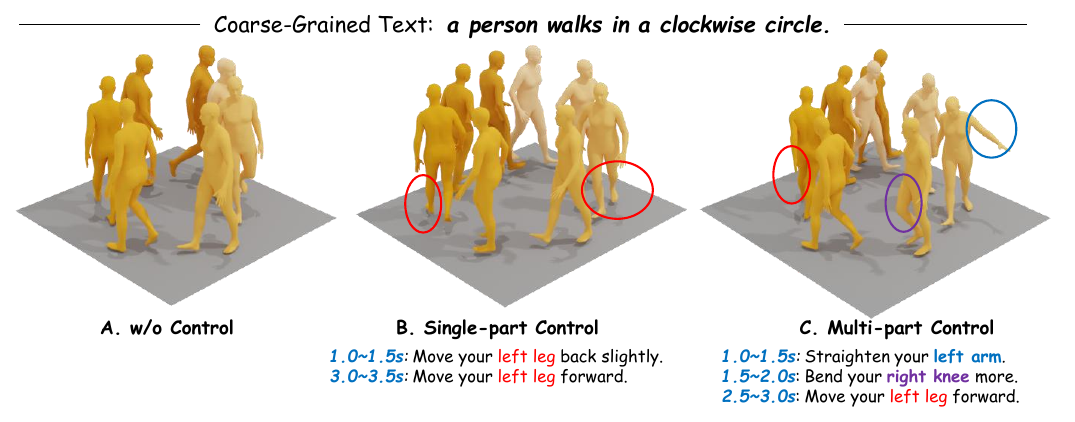}
\end{center}
\setlength{\abovecaptionskip}{-0em}  % Space above the caption
\caption{
    Visualizations of different control settings. 
    \textless Mask\textgreater\ is used for all unspecified temporal intervals.
}
% \vspace{-0.5em}
\label{fig:vis_control}
\end{figure}

\subsection{Comparison with State-of-the-arts}
\label{sec: compare with sota}
% 目的
We compare different controllable motion generation methods on HumanML3D in Tab.~\ref{table:main}.
Results in blocks (1)-(3) are taken from~\cite{OmniControl_ICLR2024,InterControl_NIPS2024}, while others are our implementations.
The implementation details can be seen in the Appendix.
From Tab.~\ref{table:main}, FineXtrol (\textit{Body Part: Average}) achieves an FID of 0.245 (\textit{vs.} 0.544) and an R-Top3 of 0.685 (\textit{vs.} 0.611), compared to MDM without control signals in block (2), which demonstrates \textbf{the effectiveness of our novel control framework}.
Compared to previous controllable motion generation methods in blocks (4)-(5), FineXtrol achieves state-of-the-art results in R-precision and Diversity, indicating its \textbf{ability to generate more diverse and accurate motions}. 
This supports the advantage of using our precise textual control signals, which offer greater flexibility than rigid spatial coordinates. Moreover, FineXtrol offers a more \textbf{user-friendly} approach to guide motion generation.
We also compare results on controlling multiple body parts, a more challenging task, in block (6).
Specifically, we follow OmniControl to train FineXtrol with combinations of body part control signals, resulting in a total of \textit{i.e.}, $\binom{6}{1} + \binom{6}{2} + \binom{6}{3} + \binom{6}{4} + \binom{6}{5} + \binom{6}{6} = 63$ combinations.
During evaluation, one combination is randomly sampled per test instance.
Results show that other methods' \textit{Cross} performance drops significantly, while FineXtrol experiences only a slight decline, highlighting the robustness of our approach.

Following ~\cite{OmniControl_ICLR2024, InterControl_NIPS2024}, we display the detailed control performance (averaged over all density levels) for six different body parts in Tab.~\ref{table:results_of_bp}. 
The results indicate that using fine-grained textual control signal of \textit{Right Arm} yields the best FID of 0.219,
using those of \textit{Right leg} achieves the highest R-Top3 of 0.689,
and using those of \textit{Left Arm} results in the best MM-Dist of 4.981.
Detailed results of different control signal densities are provided in the Appendix. 
We also display our qualitative results in Fig.~\ref{fig:vis_control}.
It shows that FineXtrol supports both single- and multi-part control over different temporal intervals.

\begin{table}[!t]
\centering
\footnotesize
\begin{tabular}{c|cccc}
\toprule
\textbf{Body Part} & \textbf{FID $\downarrow$} & \textbf{R-Top3} & \textbf{Diversity $\rightarrow$} & \textbf{MM-Dist $\downarrow$} \\

\midrule

Head  & 0.265 & 0.687 & 9.423 & 5.051 \\ 
Body & 0.261  & 0.679 & 9.548 & 5.119 \\ 
Left Arm & 0.224 & 0.684 & 9.671 & 4.981 \\ 
Right Arm & 0.219 & 0.684 & 9.427 & 5.208 \\ 
Left Leg  & 0.239 & 0.685 & 9.543 & 5.063  \\ 
Right Leg & 0.263 & 0.689 & 9.341 & 5.100 \\

\midrule

Average & \textbf{0.245} & \textbf{0.685} & \textbf{9.492} & \textbf{5.087} \\

\bottomrule

\end{tabular}
\caption{
    Detailed results of controlling specific body parts.
    }
\label{table:results_of_bp}
\end{table}

Furthermore, we report the inference time and the number of trainable parameters for our FineXtrol and other diffusion-based controllable motion generation models in Tab.~\ref{tab:inference_time}. 
Specifically, we calculated the average time required to generate a single motion sequence with 1000 denoising steps, computed over 100 runs on an NVIDIA A100-SMX-40G GPU.
The results show that FineXtrol has fewest trainable parameters, and achieves \textbf{highest inference efficiency}, benefiting from the absence of conversions between different pose representations.

\begin{table}[!t]
\footnotesize
\centering
\setlength{\tabcolsep}{3pt}
\begin{tabular}{cccc}
    \toprule[1pt]
    \multirow{2}{*}{\textbf{Methods}} & \textbf{Control} & \textbf{Inference} & \multirow{2}{*}{\textbf{Params.}}\\
      & \textbf{Signal}  & \textbf{Time(s)} $\downarrow$ & \\
    
    \midrule[0.5pt]
    % Backbone & MDM~\cite{mdm_ICLR2023}, CLIP-ViT-B/32~\cite{CLIP_ICML2021}\\
    OmniControl~\cite{OmniControl_ICLR2024}  & \multirow{3}{*}{Coord.} &168.51 & 48.79M\\
InterControl~\cite{InterControl_NIPS2024}  &  & 159.72 & 42.00M \\
GMD~\cite{GMD_CVPR2023} &  &153.25 & 238.63M\\
    % MDM & 32.35 \\
    \midrule[0.5pt]
    \textbf{FineXtrol (Ours)} & Text & \textbf{128.57} & \textbf{23.39M}
    \\
    \bottomrule[1pt]
\end{tabular}%
\caption{Inference time and the number of trainable parameters of diffusion-based controllable motion generation methods. 
`Coord.' is short for coordinate.}
% \vspace{-0.5em}
\label{tab:inference_time}
\end{table}

\begin{figure}[!t]
\begin{center}
\includegraphics[width=0.9\linewidth]{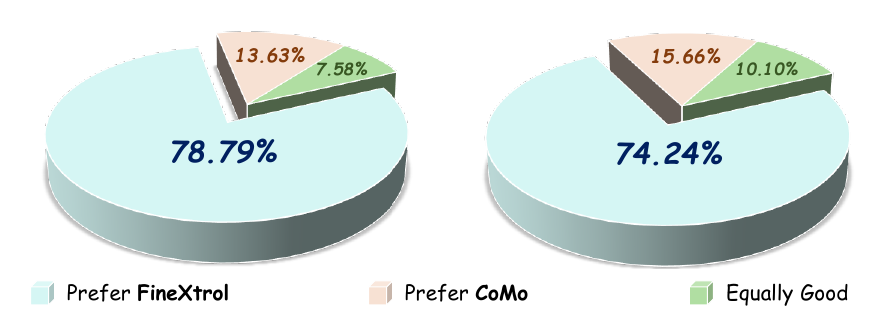}
\end{center}
\setlength{\abovecaptionskip}{-0em}  % Space above the caption
\caption{
    The statistical results of the user study. 
    The \textit{left} pie chart displays the average preference ratio for the visualized motion sequences without fine-grained textual control signals (2 cases) of our FineXtrol and CoMo. 
    The \textit{right} one shows that with fine-grained textual control signals (6 cases). 
    Each case is evaluated based on (1) alignment with control signals and (2) motion naturalness.
}
\label{fig:user_study}
\end{figure}

% UserStudy 可视化对比
\begin{figure}[!t]
\begin{center}
\includegraphics[width=1.0\linewidth]{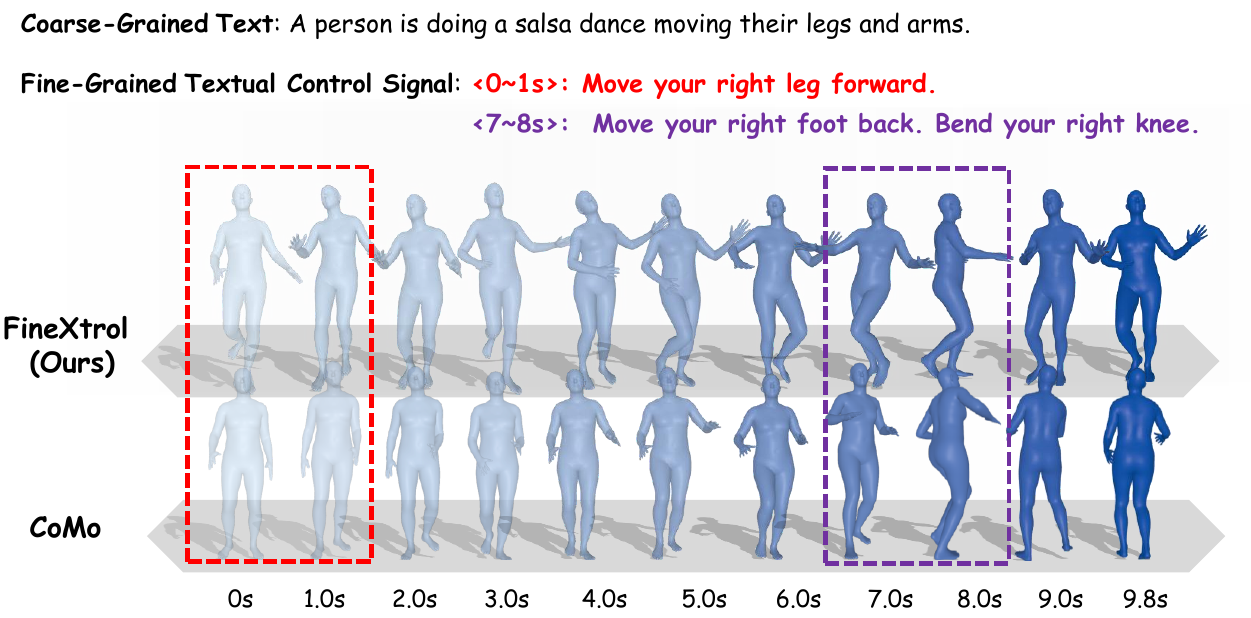}
\end{center}
\setlength{\abovecaptionskip}{-0em}  % Space above the caption
\caption{
    A motion pair comparing \textit{right leg} control in the user study.
    Body part movements in unspecified intervals are not explicitly controlled.
}
\label{fig:vis_control_user_study}
\end{figure}

We also conducted a user study with 33 subjects to compare our method with CoMo, which also uses fine-grained texts to enhance motions.
Each subject evaluated 8 cases from two perspectives.
According to the statistical results in the \textit{left} part of Fig.~\ref {fig:user_study}, our method was preferred over CoMo in 78.79\% of the cases without control signals.
With control signals (\textit{right} part of Fig.~\ref {fig:user_study}), 74.24\% of users favored our results. 
Fig.\ref{fig:vis_control_user_study} illustrates a representative example: CoMo produces motions that are misaligned with control signals, whereas FineXtrol yields more faithful and precise motion.
More comparisons are provided in the Appendix.

%%%%%%%%%%%%%%%%%%%%%%%%%%%%%%%%%%%

\subsection{Ablation Study}
\label{sec:ablation_study}
We conduct ablation experiments on both the controllable motion generation paradigm and the textual embeddings to validate the effectiveness of our framework’s design choices. 
We summarize key findings below.

\begin{table}[h]
\centering
\footnotesize
\setlength{\tabcolsep}{2.5mm} % 减少列间距
% \resizebox{\columnwidth}{!}{
\begin{tabular}{c|cc}
\toprule
Fine-grained Textual Control Paradigm & FID $\downarrow$ & R-Top3 $\uparrow$ \\

\midrule

Direct & 1.383 & 0.601  \\
\textbf{Ours} & \textbf{0.245} & \textbf{0.685} \\

\bottomrule

\end{tabular}
% }
% \vspace{-0.5em} %
\caption{
    Ablation study on control paradigm.
    Our control paradigm significantly outperforms the `Direct' paradigm, which directly connects coarse-grained text and fine-grained textual control signals as a single input.
    } 
% \vspace{-1em}
\label{table:Ablation_FG_MDM}
\end{table}

% \vspace{0.5em}
\subsubsection{Effectiveness of Our Control Paradigm.}
We directly connect the fine-grained textual control signals with the coarse-grained text, encode the combined input using our text encoder, following Fig.~\ref{fig:intro_framework}(B), and denote this baseline as `Direct'. 
% For fair comparison, we also replace our text encoder in our control paradigm with CLIP, which is used in `Direct'.
Tab.~\ref{table:Ablation_FG_MDM} indicates that `Direct' control paradigm yields poorer performance, even when the fine-grained descriptions are precise and temporally aware. 
This suggests that a single-branch model struggles to process densely packed information, and lacks the capacity to fully capture detailed textual semantics. 
In contrast, our paradigm not only preserves the ability to follow coarse-grained texts but also effectively guides the model to align with fine-grained textual control signals.

\begin{table}[!t]
\centering
\footnotesize
\setlength{\tabcolsep}{1.8mm} % 减少列间距
% \resizebox{\columnwidth}{!}{
\begin{tabular}{c|cccc}
\toprule
\textbf{Text Encoder} & \textbf{FID $\downarrow$} & \textbf{R-Top3} & \textbf{Diversity $\rightarrow$} & \textbf{MM-Dist $\downarrow$} \\

\midrule
% ~\cite{CLIP_ICML2021}
 % ~\cite{T5_JMLR2020}
CLIP  & 0.579 & 0.603 & 9.310 & 5.927 \\
T5   & 0.374 & 0.659 & 9.594 & 5.483 \\
\textbf{Ours}  & \textbf{0.245} & \textbf{0.685} & \textbf{9.492} & \textbf{5.087} \\

\bottomrule

\end{tabular}
% }
% \vspace{-0.5em} %
\caption{
    Ablation study on different text encoders. 
    The controllable motion generation performance with our text encoder significantly surpasses those with CLIP and T5, proving the effectiveness of our hierarchical contrastive learning module for fine-grained textual control signals.
    }
% \vspace{-1em}
\label{table:Motion_ablation}
\end{table}

% \vspace{0.5em}
\subsubsection{Effectiveness of the Hierarchical Contrastive Module for Fine-Grained Textual Control.} 
\begin{figure}[!t]
\begin{center}
\includegraphics[width=0.8\linewidth]{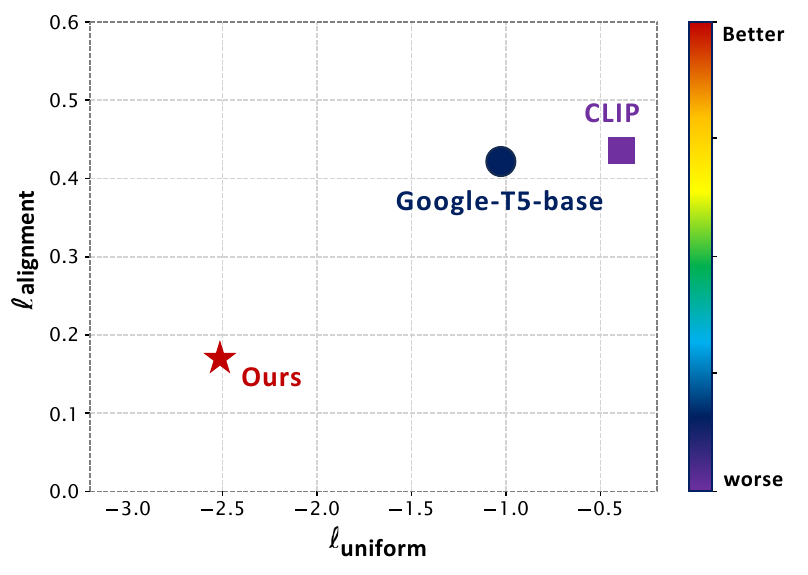}
\end{center}
\setlength{\abovecaptionskip}{-0em}  % Space above the caption
\caption{
    Comparison with existing text encoders in representing fine-grained textual control signals. 
    We plot $\ell_{\text {align}} -\ell_{\text{uniform}}$ of textual embeddings from different text encoders. 
    For both $\ell_{\text {align}}$ and $\ell_{\text{uniform}}$, lower numbers are better.
    % This scatter plot displays the comparison results with existing text encoders in representing fine-grained motion descriptions.
     % Color of points and numbers in labels represent the cosine similarity evaluated on positive pairs of sentence-level. 
    Results show that our text encoder trained with the proposed hierarchical contrastive learning module performs significantly better than prevalent text encoders.
    % The trainable copy of MDM is frozen while the fine-grained textual control signal $\boldsymbol{c}$ provides the corresponding features of the current body movement.
}
\label{fig:text_encoder_comparison}
\end{figure}

We compare different text encoders by extracting embeddings for all fine-grained textual descriptions in FineMotion in Fig.~\ref{fig:text_encoder_comparison}.
Results show that our text encoder, trained with the proposed hierarchical contrastive module, yields more discriminative embeddings than prior encoders.
More ablations for each contrastive level are provided in the Appendix.
We also assess their impact on fine-grained textual controllable motion generation in Tab.~\ref{table:Motion_ablation}.
Results show that the T5 encoder outperforms CLIP, likely due to CLIP’s truncation of long, detailed texts.
With our hierarchical contrastive module, our text encoder achieves further gains, surpassing both T5 and CLIP.

\section{Conclusion}

In this paper, we propose FineXtrol, an efficient framework that leverages novel and user-friendly control signals, \textit{i.e.}, precise and fine-grained textual descriptions of body part movements, to generate human motion sequences, offering a fresh perspective on controllable motion generation.
We further introduce a hierarchical contrastive learning module to enhance the text encoder's ability to extract discriminative embedding for this novel control signal. 
Experimental results show that FineXtrol excels in controllable motion generation, particularly in controlling body part combinations. 
Visualizations further demonstrate its ability to control specific body parts, adjust movements within specific intervals, and manipulate multiple body parts via fine-grained textual descriptions.

\section{Acknowledgments}

This work was supported by the National Key R\&D Program of China (No. 2024YFF0618403), National Natural Science Foundation of China under Grant 62576216, and Guangdong Provincial Key Laboratory under Grant 2023B1212060076.

\bibliography{aaai2026}

\newpage
\appendix

\setcounter{secnumdepth}{2}
\renewcommand{\thesubsection}{\thesection.\arabic{subsection}}
\renewcommand{\labelenumi}{\arabic{enumi}.}

% ===================================================================
% Appendix A
\section{More Details on FineXtrol}
% ===================================================================
% A.1
\subsection{Pseudo-code of Data Augmentation}
\label{appendix: data_augmentation}
% ===================================================================

This section details our data augmentation algorithms for the snippet-level and sequence-level information of the fine-grained textual control signal introduced in Sec.~\ref{sec:CL}.

The snippet-level augmentation procedure is presented in Alg.~\ref{alg:snippet_aug}.
Given a snippet-level description $\boldsymbol{c}_\text{sni}^{\text{ori}}$ from the corpus $\mathcal{D}_\text{sni}^{\text{ori}}$, we randomly replace some sentences with counterparts from $\mathcal{D}_\text{sen}^{\text{ori}}$ and shuffle their order.
The modified sentences are then connected to form the augmented snippet-level description $\boldsymbol{c}_\text{sni}^\text{aug}$.
Each snippet-level description is augmented twice to form a positive pair.

The sequence-level augmentation procedure is formalized in Alg.~\ref{alg:seq_aug}.
It takes as input a sequence-level description $\boldsymbol{c}_\text{seq}^{\text{ori}}$, which consists of an ordered list of snippet-level descriptions.
We first split $\boldsymbol{c}_\text{seq}^{\text{ori}}$ using the special token \texttt{<SEP>} into individual snippet-level descriptions, apply snippet-level augmentation to each of them, and then concatenate the augmented snippet-level descriptions in their original temporal order using \texttt{<SEP>} to reconstruct the sequence-level description.
Similarly, each sequence-level description is augmented twice to form a positive pair.

% \begin{algorithm}
% \caption{Data Augmentation for Snippet-Level}
% \label{alg:snippet_aug}
% \textbf{Input}: % Line 1030, and error occurs below:
%     \Statex $\mathcal{D}_\text{sen}^{\text{aug}}$: augmented sentence-level corpus
%     \Statex $\boldsymbol{c}_\text{sni}^\text{ori} \in \mathcal{D}_\text{sni}^{\text{ori}}$: a snippet-level description

% \textbf{Output}: An augmented snippet-level description $\boldsymbol{c}_\text{sni}^\text{aug}$
% \begin{algorithmic}[1] %[1] enables line numbers
% \Procedure{SniAug}{$\boldsymbol{c}_\text{sni}^\text{ori}$}
% \State $\ell_\text{sen} \leftarrow$ Split($\boldsymbol{c}_\text{sni}^\text{ori}$) \textcolor{black}{\hfill // \textit{Split it into a list of sentences}}
% \For{each sentence $\boldsymbol{c}_\text{sen}^j \in \ell_\text{sen}$}
%     \If{$\text{rand}() < 0.5$}
%         \State $\boldsymbol{c}_\text{sen}^j \leftarrow \mathcal{D}_\text{sen}^{\text{aug}}[\text{index\_of}(\boldsymbol{c}_\text{sen}^j)]$
%     \EndIf
% \EndFor
% \State $\ell_\text{sen} \leftarrow \text{Shuffle}(\ell_\text{sen})$  \textcolor{black}{\hfill // \textit{Randomly shuffle sentences}}
% \State $\boldsymbol{c}_\text{sni}^\text{aug} \leftarrow$ ` '.join($\ell_\text{sen}$) \textcolor{black}{\hfill // \textit{Connect sentences}}
% \State \Return $\boldsymbol{c}_\text{sni}^\text{aug}$ \textcolor{black}{\hfill // \textit{Augmented snippet-level description}}
% \EndProcedure
% \end{algorithmic}
% \end{algorithm}

\begin{algorithm}
\caption{Data Augmentation for Snippet-Level}
\label{alg:snippet_aug}

\begin{algorithmic}[1] % [1] 开启行号

\renewcommand{\algorithmicrequire}{\textbf{Input:}}
\renewcommand{\algorithmicensure}{\textbf{Output:}}

\Require $\mathcal{D}_\text{sen}^{\text{aug}}$: augmented sentence-level corpus
\Require $\boldsymbol{c}_\text{sni}^\text{ori} \in \mathcal{D}_\text{sni}^{\text{ori}}$: a snippet-level description
\Ensure An augmented snippet-level description $\boldsymbol{c}_\text{sni}^\text{aug}$

\Procedure{SniAug}{$\boldsymbol{c}_\text{sni}^\text{ori}$}
    \State $\ell_\text{sen} \leftarrow$ Split($\boldsymbol{c}_\text{sni}^\text{ori}$) \textcolor{black}{\hfill // \textit{Split it into a list of sentences}}
    \For{each sentence $\boldsymbol{c}_\text{sen}^j \in \ell_\text{sen}$}
        \If{$\text{rand}() < 0.5$}
            \State $\boldsymbol{c}_\text{sen}^j \leftarrow \mathcal{D}_\text{sen}^{\text{aug}}[\text{index\_of}(\boldsymbol{c}_\text{sen}^j)]$
        \EndIf
    \EndFor
    \State $\ell_\text{sen} \leftarrow \text{Shuffle}(\ell_\text{sen})$  \textcolor{black}{\hfill // \textit{Randomly shuffle sentences}}
    \State $\boldsymbol{c}_\text{sni}^\text{aug} \leftarrow$ ` '.join($\ell_\text{sen}$) \textcolor{black}{\hfill // \textit{Connect sentences}}
    \State \Return $\boldsymbol{c}_\text{sni}^\text{aug}$ \textcolor{black}{\hfill // \textit{Augmented snippet-level description}}
\EndProcedure

\end{algorithmic}
\end{algorithm}

\begin{algorithm}
\caption{Data Augmentation for Sequence-Level}
\label{alg:seq_aug}
\begin{algorithmic}[1] % [1] 开启行号

    \renewcommand{\algorithmicrequire}{\textbf{Input:}}
    \renewcommand{\algorithmicensure}{\textbf{Output:}}

    \Require $\boldsymbol{c}_\text{seq}^\text{ori}$: a sequence-level description
    \Ensure An augmented sequence-level description $c_\text{sni}^\text{aug}$

    \Procedure{SeqAug}{$\boldsymbol{c}_\text{seq}^\text{ori}$}
        \State $\ell_\text{sni} \leftarrow \text{Split}(\boldsymbol{c}_\text{seq}^\text{ori})$ \textcolor{black}{\hfill // \textit{Split it into a list of snippets}}
        \For{each snippet $c_\text{sni}^{i} \in \ell_\text{sni}$}
            \State $c_\text{sni}^{i} \leftarrow \text{SNIAUG} (c_\text{sni}^{i})$ \textcolor{black}{\hfill // \textit{Algorithm~\ref{alg:snippet_aug}}}
        \EndFor
        \State $c_\text{sni}^\text{aug} \leftarrow $ ` $\textless \text{SEP} \textgreater$ '.join($\ell_\text{sni}$)  \textcolor{black}{\hfill // \textit{Connect snippets}}
        \State \Return $c_\text{seq}^\text{aug}$ \textcolor{black}{\hfill // \textit{Augmented sequence description}}
    \EndProcedure

\end{algorithmic}
\end{algorithm}

% A.1 ends here
% ===================================================================
% A.2 
\subsection{Prompt for paraphrasing}
\label{appendix: ds_prompt}
% ===================================================================

\begin{figure}[h]
    \centering
    \includegraphics[
        % Sets the height to 30% of the column's text height.
        % Width will adjust automatically.
        height=0.3\textheight,
        width=0.95\linewidth,  % Keep your desired width
        keepaspectratio,      % This is the crucial option
    ]{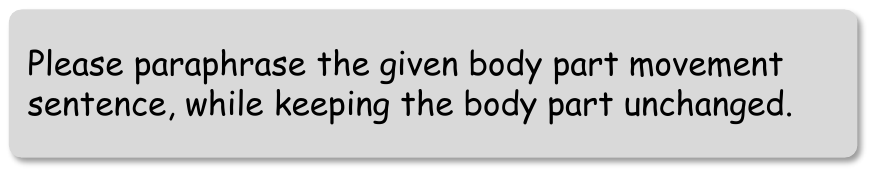}
    \caption{The prompt for paraphrasing sentence-level descriptions.}
    \label{fig:appendix_sent_prompt}
\end{figure}

In Sec.~\ref{sec:sentence-level}, we construct an augmented sentence-level corpus $\mathcal{D}_\text{sen}^{\text{aug}}$.
Specifically, we employ the large language model, DeepSeek-V2~\cite{deepseekv2}, to paraphrase each sentence in the original corpus $\mathcal{D}_\text{sen}^{\text{ori}}$. 
Here, we display the prompt used for this paraphrasing task is shown in Fig.~\ref{fig:appendix_sent_prompt}.

% {
%     \centering
%     \includegraphics[
%         trim=0pt 5pt 0pt 0pt, 
%         clip,                    
%         width=0.9\linewidth     
%     ]{img/0801_sent_para.pdf}
%      % \setlength{\abovecaptionskip}{0.2em} % 裁剪后可以适当增加一点标题和图片的距离
%     \captionof{figure}{
%        The prompt provided to DeepSeek for generating sentence paraphrases.
%     }
%     \label{fig:appendix_sent_prompt}
%     \par
% }

% A.2 ends here
% ===================================================================
% ===================================================================

% A.3 
\subsection{Detail of Body Parts Division}
\label{appendix: body_division}
% ===================================================================
% Following the evaluating benchmarks of OmniControl ~\cite{OmniControl_ICLR2024} and the structure of the fine-grained textual description from FineMotion ~\cite{Wu_FineMotion_ICCV2025}, we propose an appropriate and novel division for generating controllable motion sequences as mentioned in Sec ~\ref{sec:Background}. Tab ~\ref{table:body_part_division} shows the details of our division strategy. We preliminarily divide all the body parts into \textbf{6} key parts and each body part contains at least \textbf{1} joint in detail. During training and evaluating, we mask a target body part at a specific ratio, which means all the sentences containing the related descriptions of this body part will be replaced by the special tokens \texttt{<Mask>}.

OmniControl~\cite{OmniControl_ICLR2024} considers six key joints—pelvis, left foot, right foot, head, left wrist, and right wrist—based on their frequent involvement in interactions with objects and the surrounding environment. 
These joints effectively divide the human body into six main parts.
To align with this protocol, we analyze all descriptions about body parts mentioned in fine-grained annotations of FineMotion~\cite{Wu_FineMotion_ICCV2025} and observe that the descriptions in FineXtrol feature more diverse references: some descriptions target individual joints, while others involve multiple joints.
For consistency in evaluation, we also group body parts into six main parts, following the same division strategy.
These main parts, along with their corresponding descriptions, are listed in Tab.~\ref{table:body_part_division}.

% and group them into six main parts.

\begin{table}[H]
\centering
\footnotesize
\begin{tabularx}{\columnwidth}{p{2cm}X}
\toprule
\rowcolor{white} % 表头保持白色
\textbf{Body Parts} & \textbf{Related Descriptions} \\
\midrule
\multirow{1}{*}{Head} & head \\
\midrule
\multirow{1}{*}{Body} & body, torso, waist, upper back, lower back \\
\midrule
\multirow{3}{*}{Left Arm} & left hand, left arm, left elbow, left shoulder, left forearm, hands, arms, elbows, shoulders, forearms \\
\midrule
\multirow{3}{*}{Right Arm} & right hand, right arm, right elbow, right shoulder, right forearm, hands, arms, elbows, shoulders, forearms \\
\midrule
\multirow{2}{*}{Left Leg} & left leg, left foot, left knee, left heel, legs, feet, knees, heels \\
\midrule
\multirow{2}{*}{Right Leg} & right leg, right foot, right knee, right heel, legs, feet, knees, heels \\
\bottomrule
\end{tabularx}
\caption{Controllable body parts and related descriptions.}
\label{table:body_part_division}
\end{table}

% A.3 ends here
% ===================================================================
% A.4 
\onecolumn

\subsection{Data Augmentation Examples for Three Levels of Information in Fine-Grained Textual Control Signals}
\label{appendix: dt_example}
% ===================================================================

As introduced in Sec.\ref{sec:CL}, the fine-grained control signal $\boldsymbol{c}$ comprises three levels of information.
The data augmentation strategies for each level are detailed in Sec.\ref{sec:CL} and Sec.~\ref{appendix: data_augmentation}.
Here, we display examples of three levels of descriptions along with their augmented version in the table below.

\begin{table*}[!h]
\centering
\small

\begin{tabular}{p{8cm} p{8cm}}
\toprule

\textbf{Original Description} & \textbf{Augmented Description}\\
\midrule

\textbf{Sentence-level}\\
\midrule
\textbf{\#$\text{SEN}_1$:} Bend your body forward and put your hands in front of your thighs. 
& \textbf{\#$\text{SEN}_1^{'}$:} Lean forward and put your hands in front of your thighs. \\
\textbf{\#$\text{SEN}_2$:} Gently bring your left leg down to match the position of your right leg.
& \textbf{\#$\text{SEN}_2^{'}$:} Lower your left leg so it is in the same position as your right leg. \\
\textbf{\#$\text{SEN}_3$:} Straighten your left leg and put your left foot on the floor.
& \textbf{\#$\text{SEN}_3^{'}$:} Extend your left leg and place your left foot flat on the ground. \\
\textbf{\#$\text{SEN}_4$:} Move your left arm closer to your body.
& \textbf{\#$\text{SEN}_4^{'}$:} Bring your left arm nearer to your torso.\\
\textbf{\#$\text{SEN}_5$:} Bring your right arm up a little.
& \textbf{\#$\text{SEN}_5^{'}$:} Lift your right arm slightly. \\
% \textbf{\#$\text{SEN}_6$:} Move your left elbow down to your left hip. 
% & \textbf{\#$\text{SEN}_6^{'}$:} Lower your left elbow towards your left hip.  \\
\midrule

\textbf{Snippet-level}\\
\midrule
\textbf{\#$\text{SNI}_1$:} Extend your right arm backward, positioning it behind your body. Bring your right arm down slightly. Gently bring your right hand down to match the position of your right foot.
& \textbf{\#$\text{SNI}_1^{'}$:} Lower your right hand so it is in the same position as your right foot. Bring your right arm back behind you. Bring your right arm down slightly. \\
\textbf{\#$\text{SNI}_2$:} Straighten your left leg. Put your left foot on the floor. 
% Move your right arm to the right. Move your left hand to your left thigh.
& \textbf{\#$\text{SNI}_2^{'}$:} Put your left foot on the floor. Straighten your left leg. \\
% and Move your right arm to the right. Gently transfer your left hand to rest upon your left thigh. \\
\textbf{\#$\text{SNI}_3$:} Move your left leg forward and bend your left knee slightly. Move your left leg back slightly. 
% Move your left hand to your left thigh. Move your right hand to your right thigh.
& \textbf{\#$\text{SNI}_3^{'}$:} Move your left leg back slightly.  Move your left leg forward and bend your left knee.  \\
% Move your left hand to your left thigh. Extend your right hand towards your right thigh.
\textbf{\#$\text{SNI}_4$:}Move your left hand to your right thigh. Gently transfer your left hand to rest upon your left thigh. 
& \textbf{\#$\text{SNI}_4^{'}$:} Move your left hand to your left thigh. Move your left hand to your right thigh.  
\\
\textbf{\#$\text{SNI}_5$:} Move your right leg forward and point your right foot on the floor. Move your right leg back and point your right foot on the floor. 
& \textbf{\#$\text{SNI}_5^{'}$:} Step back with your right leg, ensuring your right foot is flat on the ground. Move your right leg forward and point your right foot on the floor.
\\
\midrule

\textbf{Sequence-level}\\
\midrule
\textbf{\#$\text{SEQ}_1$:} \newcommand{\mytoken}[1]{\textless #1\textgreater\allowbreak}
\mytoken{Motionless} \mytoken{SEP} Move your left leg back slightly. \mytoken{SEP} Move your left foot back slightly. \mytoken{SEP} Move your left leg forward. \mytoken{SEP} \mytoken{Motionless} \mytoken{SEP} \mytoken{Motionless} \mytoken{SEP} \mytoken{Motionless} \mytoken{SEP} \mytoken{Motionless} \mytoken{SEP} \mytoken{Motionless} \mytoken{SEP} Move your left leg back. \mytoken{SEP} \mytoken{Motionless} \mytoken{SEP} \mytoken{Motionless} \mytoken{SEP} \mytoken{Motionless} \mytoken{SEP} \mytoken{Motionless} \mytoken{SEP} \mytoken{Motionless} \mytoken{SEP} \mytoken{Motionless} \mytoken{SEP} \mytoken{Motionless}

& \textbf{\#$\text{SEQ}_1^{'}$:} \newcommand{\mytoken}[1]{\textless #1\textgreater\allowbreak}
\mytoken{Motionless} \mytoken{SEP} Slightly draw your left leg backward. \mytoken{SEP} Shift your left foot back a bit. \mytoken{SEP} Bring your left leg forward. \mytoken{SEP} \mytoken{Motionless} \mytoken{SEP} \mytoken{Motionless} \mytoken{SEP} \mytoken{Motionless} \mytoken{SEP} \mytoken{Motionless} \mytoken{SEP} \mytoken{Motionless} \mytoken{SEP} Pull your left leg backward. \mytoken{SEP} \mytoken{Motionless} \mytoken{SEP} \mytoken{Motionless} \mytoken{SEP} \mytoken{Motionless} \mytoken{SEP} \mytoken{Motionless} \mytoken{SEP} \mytoken{Motionless} \mytoken{SEP} \mytoken{Motionless} \mytoken{SEP} \mytoken{Motionless}\\ 

\textbf{\#$\text{SEQ}_2$:} \newcommand{\mytoken}[1]{\textless #1\textgreater\allowbreak}
Move your left hand to your left thigh. \mytoken{SEP} \mytoken{Motionless} \mytoken{SEP} \mytoken{Motionless} \mytoken{SEP} \mytoken{Motionless} \mytoken{SEP} \mytoken{Motionless} \mytoken{SEP} \mytoken{Motionless} \mytoken{SEP} Move your left hand to your left thigh. \mytoken{SEP} \mytoken{Motionless} 

& \textbf{\#$\text{SEQ}_2^{'}$:} \newcommand{\mytoken}[1]{\textless #1\textgreater\allowbreak}
Place your left hand on your left thigh. \mytoken{SEP} \mytoken{Motionless} \mytoken{SEP} \mytoken{Motionless} \mytoken{SEP} \mytoken{Motionless} \mytoken{SEP} \mytoken{Motionless} \mytoken{SEP} \mytoken{Motionless} \mytoken{SEP} Place your left hand on your left thigh. \mytoken{SEP} \mytoken{Motionless} \\ 

\textbf{\#$\text{SEQ}_3$:} \newcommand{\mytoken}[1]{\textless #1\textgreater\allowbreak}
Lean your upper body to the right. \mytoken{SEP} \mytoken{Motionless} \mytoken{SEP} \mytoken{Motionless} \mytoken{SEP} Lean your body to the right. \mytoken{SEP} Bend your body a little more to the right. Lower your hands to your waist. \mytoken{SEP} \mytoken{Motionless} \mytoken{SEP} \mytoken{Motionless} \mytoken{SEP} Lean your upper body to the right. Bring your left elbow down to your waist.  \mytoken{SEP} Lean your upper body to the left. Bring your left elbow down to your waist. \mytoken{SEP} \mytoken{Motionless} \mytoken{SEP} \mytoken{Motionless} \mytoken{SEP} \mytoken{Motionless} \mytoken{SEP} Lean your body to the right. \mytoken{SEP} \mytoken{Motionless} \mytoken{SEP} \mytoken{Motionless}

& \textbf{\#$\text{SEQ}_3^{'}$:} \newcommand{\mytoken}[1]{\textless #1\textgreater\allowbreak}
Tilt your upper body to the right. \mytoken{SEP} \mytoken{Motionless} \mytoken{SEP} \mytoken{Motionless} \mytoken{SEP} Tilt your body to the right. \mytoken{SEP} Incline your body slightly further to the right. Bring your hands down to your waist. \mytoken{SEP} \mytoken{Motionless} \mytoken{SEP} \mytoken{Motionless} \mytoken{SEP} Tilt your upper body to the right. Lower your left elbow to your waist. \mytoken{SEP} Tilt your upper body to the left. Lower your left elbow to your waist. \mytoken{SEP} \mytoken{Motionless} \mytoken{SEP} \mytoken{Motionless} \mytoken{SEP} \mytoken{Motionless} \mytoken{SEP} Tilt your body to the right. \mytoken{SEP} \mytoken{Motionless} \mytoken{SEP} \mytoken{Motionless}\\ 
\bottomrule
\end{tabular}
% \vspace{-2mm} %
\caption{
Examples of data augmentation for three levels of information in fine-grained textual control signals.
}
\label{tab:retrieved-examples}
\end{table*}

\twocolumn
% A.4 ends here
% ===================================================================
% Appendix B
\section{More Experimental Details and Results}
% ===================================================================
% B.1 
\subsection{Details of Datasets}
\label{appendix: dataset}
% ===================================================================

% We conduct experiments on the widely used \textbf{HumanML3D}~\cite{T2M_CVPR2022} dataset, which comprises \textbf{14,616} motion sequences from \textbf{AMASS}~\cite{Amass_ICCV2019} and \textbf{HumanAct12}~\cite{Action2motion_ACMMM2020}, along with textbf{44,970} textual descriptions. 
% These textual descriptions serve as the coarse-grained motion descriptions $\boldsymbol{p}$.
% As for the fine-grained textual control signals $\boldsymbol{c}$, we use the open-source \textbf{FineMotion}~\cite{Wu_FineMotion_ICCV2025} dataset, which extends HumanML3D by providing fine-grained annotations for textbf{420,968} short motion snippets.

We conduct experiments on the widely-used \textbf{HumanML3D}~\cite{T2M_CVPR2022} dataset, which comprises 14,616 motion sequences from AMASS~\cite{Amass_ICCV2019} and HumanAct12~\cite{Action2motion_ACMMM2020}, along with 44,970 textual descriptions. 
These textual descriptions serve as the coarse-grained text $\boldsymbol{p}$.

For the fine-grained control signals $\boldsymbol{c}$, we use the \textbf{FineMotion}~\cite{Wu_FineMotion_ICCV2025} dataset, which extends HumanML3D by providing fine-grained textual descriptions for all body parts across \textbf{420,968} short motion snippets.
We extract descriptions of specific body parts to serve as the fine-grained control signals.

% B.1 ends here
% ===================================================================
% B.2 
\subsection{More Implementation Details}
\label{appendix: implementation details}
% ===================================================================
\subsubsection{Model details.} 
As shown in Fig.\ref{fig:model_overview}, our model comprises two branches that are fine-tuned jointly.
Each branch contains 8 layers of Transformer encoders with a latent dimension of 512.
The entire framework has a total of 183M parameters.
The encoded feature of our fine-grained textual control signals, $\boldsymbol{e}_c$, has a dimensionality of $C=768$.

% \vspace{0.5em}
\subsubsection{Training details.} 
The training process consists of two stages.
In the first stage, we train the text encoder using a hierarchical contrastive learning module to encode fine-grained textual control signals into discriminative embeddings.
In the second stage, we train the entire FineXtrol framework, initializing the text encoder with the weights obtained from the previous stage and keeping it frozen throughout.

When training the text encoder, we adopt T5~\cite{T5_JMLR2020} as the base text encoder due to its capacity to encode long text sequences.
The encoder is progressively trained using contrastive learning at the sentence, snippet, and sequence levels, with each level initialized from the weights learned at the previous one.
The same set of hyperparameters is used across all three levels. 
Specifically, we use the AdamW optimizer~\cite{AdamW_ICLR2019} and a batch size of 64. 
Notably, for long sequence-level inputs, we use a batch size of 16 with gradient accumulation over 4 steps to maintain an effective batch size of 64.
The learning rate follows a linear warm-up for the first 1500 iterations, remains constant at $1\times10^{-5}$, and then decays following a cosine annealing schedule. 
Each level of contrastive learning is trained for 200 epochs.

For training FineXtrol, we follow the hyperparameter settings of OmniControl~\cite{OmniControl_ICLR2024}. We use the AdamW optimizer with a learning rate of $1\times10^{-5}$ and a batch size of 64. adopting the total number of training iterations, which is set to 400K (approximately 40 hours on a single A100 40G GPU).

Our framework is trained with 1000 diffusion steps and a cosine noise schedule.
Following the classifier-free guidance technique~\cite{Classifier_free}, we randomly mask 10\% of the coarse-grained texts $\boldsymbol{p}$ during training.
To encode the control signal $\boldsymbol{c}$, as described in Sec.~\ref{sec:Motion ControlNet}, we randomly replace descriptions within temporal intervals with the special token \texttt{<Mask>} at a probability of 50\%.

\subsection{Comparison with CoMo}
\label{appendix:vs_CoMo}

We note that Como~\cite{CoMo_ECCV2024} also incorporates fine-grained textual descriptions to generate precise motion sequences.
Specifically, it defines 10 different body parts (head, torso, left/right arm, left/right hand, left/right leg, left/right foot) plus one additional category for overall mood.
Then, it leverages LLMs to generate detailed descriptions for each part.
These descriptions are then encoded using the CLIP~\cite{CLIP_ICML2021} text encoder to produce an $11 \times 512$ feature embedding for each motion sequence: $10 \times 512$ for the detailed body parts and $1 \times 512$ for the mood embedding.
The embeddings are flattened and concatenated with the embedding of the coarse-grained text before being fed into the motion generation model.

However, as discussed in Sec.~\ref{sec:intro} of our main paper, its LLM-generated descriptions suffer from two key limitations: 
(1) they are not always aligned with the ground-truth motion, and 
(2) they lack explicit temporal cues, resulting in weak correspondence between the text and motion segments. 
These issues limit the controllability over both joint movements and temporal intervals. 
To highlight the strengths of our fine-grained control signals, we conduct extensive qualitative and quantitative experiments.

For fair comparisons, we first map CoMo’s 10 body parts to the 6 body parts used in our control framework, as detailed in Tab.~\ref{table:como_align}.

\begin{table}[H]
\centering
\footnotesize
\setlength{\tabcolsep}{16pt}
% 将 tabularx 换成 tabular，并保留 ccc (居中) 定义
\begin{tabular}{ll}
\toprule
% \rowcolor{white} 
\textbf{Body Parts in FineXtrol} & \textbf{Body Parts in CoMo} \\
\midrule
% 移除了不必要的 \multirow{1}
Head & head  \\
Torso & Torso  \\
Left Arm & L-Arm, L-Hand \\
Right Arm & R-Arm, R-Hand \\
Left Leg & L-Leg, L-Feet \\
Right Leg & R-Leg, R-Feet \\
\bottomrule
\end{tabular}
\caption{Conversions of body parts in FineXtrol and CoMo.}
\label{table:como_align}
\end{table}

\subsubsection{Quantitative Comparison with CoMo.} 
\label{sec:COMO_Quantitative_SM}

Notably, CoMo focuses on precise motion generation rather than controllable motion generation.
To evaluate its controllability, we set the textual embeddings of uncontrolled body parts to zero vectors and randomly replace textual embeddings of uncontrolled parts with zero vectors at probabilities of 0\%, 25\%, 50\%, and 75\%, aligning with our masking strategy for fine-grained textual control signals.
This alignment allows us to quantitatively assess CoMo’s controllable generation performance on each of our six defined body parts, providing a direct and fair benchmark for comparison.

\begin{table*}[!t]
\begin{center}
\footnotesize
\setlength{\tabcolsep}{2pt}
\begin{tabular*}{0.95\textwidth}{@{\extracolsep{\fill}}cc c c cc}
    \toprule
    
    \textbf{Density of Control Signal} & \textbf{Body Part} & \textbf{FID $\downarrow$} & \textbf{R-precision $\uparrow$ (Top-3)} & \textbf{Diversity $\rightarrow$ 9.503} & \textbf{MM-Dist $\downarrow$} \\

    \midrule

    25\% & \multirow{4}{*}{Head} & 0.354 & 0.625 & 9.987  & 5.915 \\
    50\% &                       & 0.351 & 0.616 & 9.591  & 5.811 \\
    75\% &                       & 0.328 & 0.625 & 9.405  & 5.749 \\
    100\%  &                     & 0.360 & 0.632 & 9.339  & 5.694 \\
    \midrule
    25\% & \multirow{4}{*}{Body} & 0.363 & 0.625 & 9.586  & 5.905 \\
    50\% &                       & 0.382 & 0.613 & 9.213  & 5.894 \\
    75\% &                       & 0.331 & 0.631 & 9.603  & 5.530 \\
    100\%  &                     & 0.345 & 0.621 & 9.253  & 5.501 \\
    \midrule
    25\% & \multirow{4}{*}{Left Arm} & 0.371 & 0.627 & 9.732  & 5.505 \\
    50\% &                           & 0.367 & 0.619 & 9.447  & 5.506 \\
    75\% &                           & 0.335 & 0.633 & 9.345  & 5.477 \\
    100\%  &                         & 0.347 & 0.635 & 9.416  & 5.484 \\
    \midrule
    25\% & \multirow{4}{*}{Right Arm} & 0.351 & 0.627 & 9.847  & 5.574 \\
    50\% &                            & 0.352 & 0.615 & 9.854  & 5.530 \\
    75\% &                            & 0.311 & 0.623 & 9.801  & 5.494 \\
    100\%  &                          & 0.306 & 0.633 & 9.502  & 5.469 \\
    \midrule
    25\% & \multirow{4}{*}{Left Leg} & 0.356 & 0.622 & 9.713  & 5.517 \\
    50\% &                           & 0.355 & 0.620 & 9.387  & 5.515 \\
    75\% &                           & 0.324 & 0.626 & 9.533  & 5.495 \\
    100\%  &                         & 0.345 & 0.632 & 9.657  & 5.512 \\
    \midrule
    25\% & \multirow{4}{*}{Right Leg} & 0.367 & 0.617 & 9.636  & 5.533 \\
    50\% &                            & 0.349 & 0.620 & 9.671  & 5.524 \\
    75\% &                            & 0.331 & 0.633 & 9.577  & 5.480 \\
    100\%  &                          & 0.348 & 0.629 & 9.534  & 5.488 \\

    % \midrule
    % Avg.  & - & 0.347 & 0.625 & 9.568  & 5.588  \\

    \bottomrule
    
\end{tabular*}
\caption{
    The detailed controllable motion generation results of CoMo on HumanML3D.
} \label{table:eval_como}
\end{center}
\end{table*}

In Sec.~\ref{sec: compare with sota}, we report the average performance of controlling different body parts under four levels of control signal density for FineXtrol and CoMo in Tab.\ref{table:main} in the main paper.
Results show that our FineXtrol demonstrates a notable advantage in motion quality and text-motion alignment.
These results suggest that CoMo is unable to provide precise control signals to a single body part and its performance degrades when control is isolated to specific body parts via masking.
Here, we further provide the detailed results of CoMo for each body part across the four levels of control density in Tab.~\ref{table:eval_como}.

\begin{figure}[!t]
\begin{center}
\includegraphics[width=1.0\columnwidth]{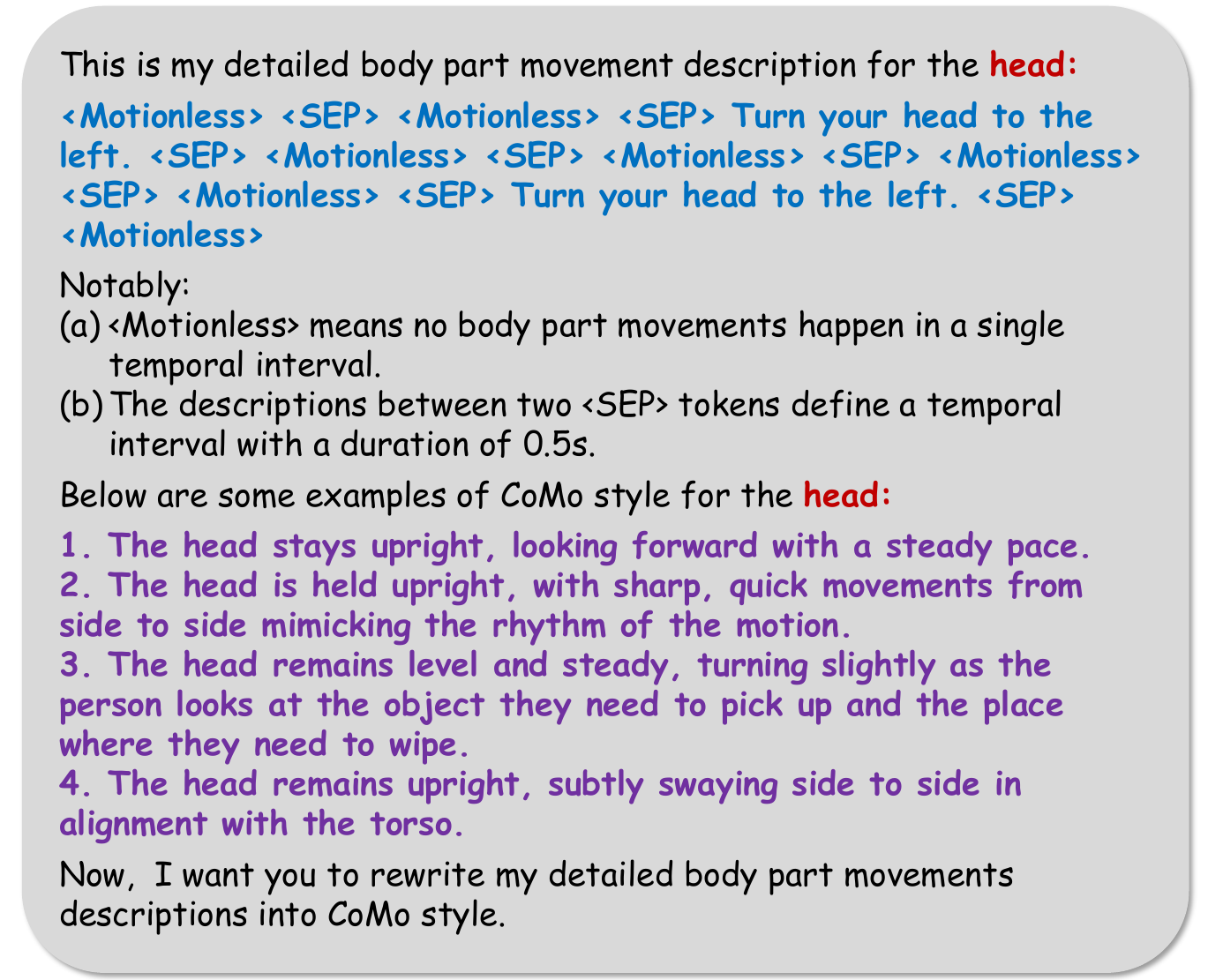}
\end{center}
\setlength{\abovecaptionskip}{-0em}  % Space above the caption
\caption{
    Prompting template for converting our fine-grained textual control signals into the fine-grained description style used in CoMo.
}
\label{fig:appendix_como_prompt}
\end{figure}

\begin{figure}[!t]
\begin{center}
\includegraphics[width=1.0\linewidth]{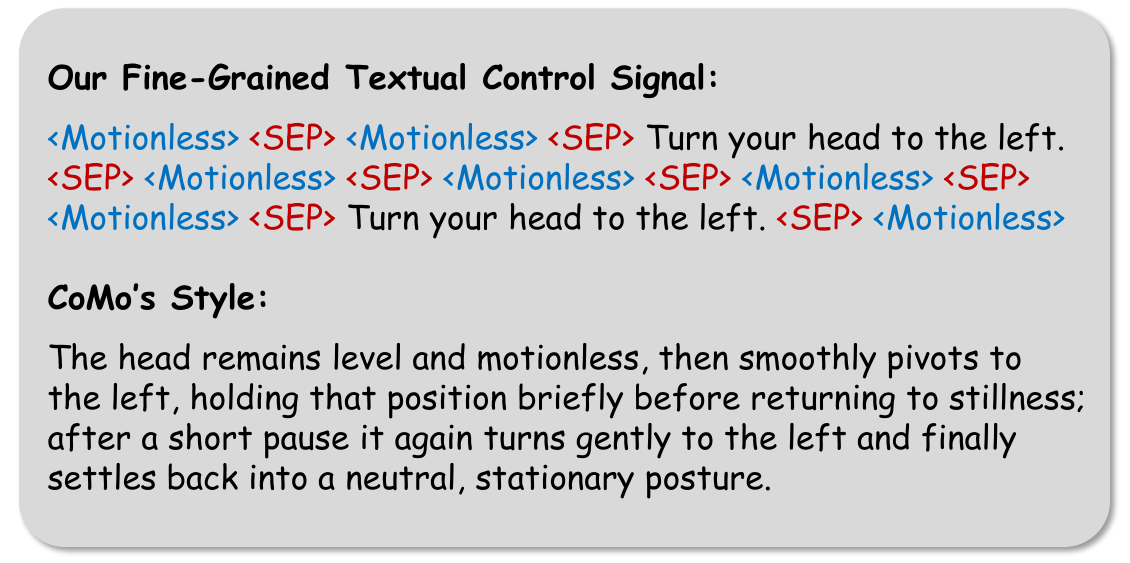}
\end{center}
\setlength{\abovecaptionskip}{-0em}  % Space above the caption
\caption{
    An example of rewritten fine-grained texutal control signal in CoMo's style.
}
\label{fig:appendix_como_dt}
\end{figure}

\begin{figure*}[!h]
\begin{center}
\includegraphics[width=1.0\linewidth]{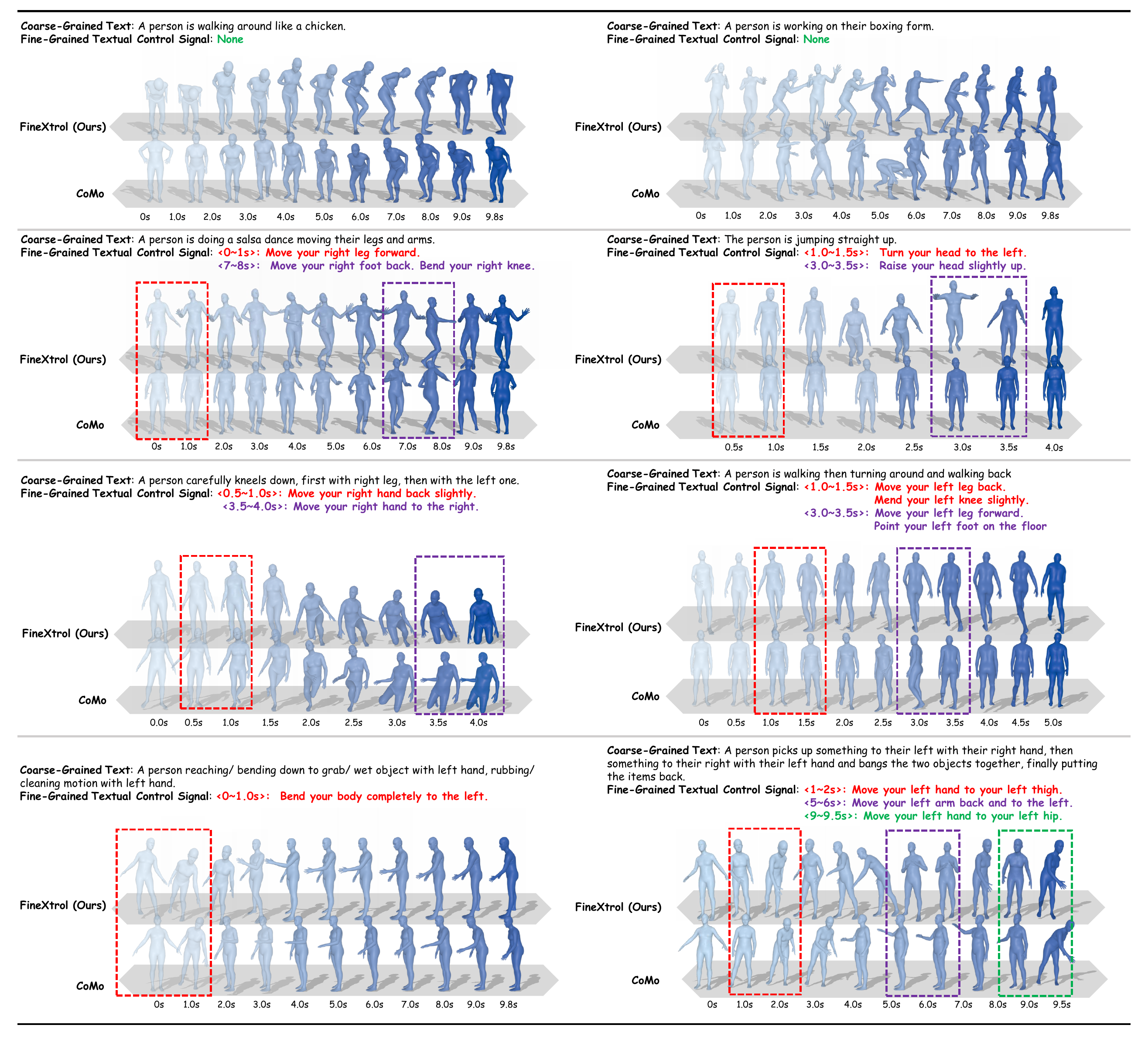}
\end{center}
\setlength{\abovecaptionskip}{-0em}  % Space above the caption
\caption{
  More qualitative comparisons with CoMo.
  Body part movements in unspecified intervals are not explicitly controlled.
}
\label{fig:userstudy_vis}
\end{figure*}

% The quantitative comparison results are presented in Table~\ref{table:eval_como}. CoMo achieves competitive results, with an average FID of 0.3471, an Top-3 R-Precision of 0.6250, and a Diversity score (9.5678) comparable to the ground truth (\textit{vs.}~9.503). 
% Despite these strong \textcolor{red}{results}, our FineXtrol demonstrates a notable advantage in motion quality (FID: 0.249) and text-motion alignment (R-Precision: 0.684; Matching Score: 5.114). 
% \textcolor{red}{These results suggest that CoMo is unable to provide precise control signals to single body part and its performance degrades when control is isolated to specific body parts via masking.}

% xThese results suggest that CoMo's performance degrades when control is isolated to specific body parts via masking, highlighting the superior robustness and precision of our framework.

\subsubsection{Qualitative Comparison with CoMo.} 
We displayed more qualitative comparisons with CoMo—also used in our user study (see Sec.\ref{appendix: User Study})—focusing on the control of body part movements within specific temporal intervals, as shown in Fig.\ref{fig:userstudy_vis}.
As we mentioned before, the LLM-augmented fine-grained descriptions lack explicit temporal cues.
To ensure a fair comparison, we rephrase our fine-grained textual control signals to match CoMo’s descriptive style.
This is achieved using GPT-4o with a carefully crafted prompt, shown in Fig.~\ref{fig:appendix_como_prompt}.
The prompt includes several example descriptions from CoMo to guide the LLM in aligning our control signals with CoMo’s format.

Fig.\ref{fig:appendix_como_dt} shows an example of a rewritten fine-grained textual control signal.
We use the CLIP\cite{CLIP_ICML2021} text encoder to extract textual embeddings from the rewritten descriptions corresponding to the controlled body part, while setting the embeddings of uncontrolled body parts to zero vectors.
These embeddings are then used to generate controlled motions with CoMo.
This setup ensures that the resulting visualizations provide a fair comparison and accurately reflect each model's capabilities under identical control conditions.

\begin{table*}[!t]
\centering
\small
\setlength{\tabcolsep}{10pt}
\resizebox{0.95\textwidth}{!}{
\begin{tabular}{cccccc}
\toprule
\textbf{Density of Control Signal} & \textbf{Body Part} & \textbf{FID $\downarrow$} & \textbf{R-precision $\uparrow$ (Top-3)} & \textbf{Diversity $\rightarrow$ 9.503} & \textbf{MM-Dist $\downarrow$} \\
\midrule
25\% & Head & 0.238 & 0.686 & 9.424 & 4.925 \\
50\% & Head & 0.265 & 0.683 & 9.665 & 4.993 \\
75\% & Head & 0.297 & 0.682 & 9.005 & 5.119 \\
100\% & Head & 0.259 & 0.700 & 9.597 & 5.165 \\
\midrule
25\% & Body & 0.254 & 0.649 & 9.123 & 5.193 \\
50\% & Body & 0.239 & 0.678 & 9.706 & 5.108 \\
75\% & Body & 0.238 & 0.699 & 9.900 & 5.009 \\
100\% & Body & 0.314 & 0.690 & 9.463 & 5.166 \\
\midrule
25\% & Left Arm & 0.305 & 0.685 & 9.512 & 4.823 \\
50\% & Left Arm & 0.183 & 0.697 & 9.681 & 4.895 \\
75\% & Left Arm & 0.207 & 0.659 & 9.931 & 5.093 \\
100\% & Left Arm & 0.203 & 0.696 & 9.559 & 5.111 \\
\midrule
25\% & Right Arm & 0.213 & 0.689 & 9.592 & 5.206 \\
50\% & Right Arm & 0.226 & 0.654 & 9.104 & 5.200 \\
75\% & Right Arm & 0.239 & 0.672 & 9.534 & 5.224 \\
100\% & Right Arm & 0.199 & 0.720 & 9.477 & 5.202 \\
\midrule
25\% & Left Leg & 0.292 & 0.672 & 9.640 & 5.171 \\
50\% & Left Leg & 0.181 & 0.677 & 9.673 & 5.020 \\
75\% & Left Leg & 0.289 & 0.698 & 9.305 & 4.938 \\
100\% & Left Leg & 0.194 & 0.692 & 9.552 & 5.123 \\
\midrule
25\% & Right Leg & 0.258 & 0.694 & 9.199 & 5.163 \\
50\% & Right Leg & 0.250 & 0.682 & 9.500 & 5.029 \\
75\% & Right Leg & 0.271 & 0.696 & 9.170 & 5.013 \\
100\% & Right Leg & 0.273 & 0.696 & 9.496 & 5.195 \\
% \midrule
% Avg. & - & 0.249 &	0.684 &	9.511 & 5.113 \\

% Avg. & - & 0.249 &	0.684&	9.511 & 5.113 \\

\bottomrule

\end{tabular}
}
\caption{
The detailed results of our FineXtrol on HumanML3D.
% The table displays the complete performance metrics (FID, R-precision, Diversity, and Matching Score) for different ratios of masked sentences on six different body parts of the proposed method.
% $\downarrow$ indicates lower values are better, $\uparrow$ indicates higher values are better, and $\rightarrow$ indicates values closer to real data are better.
}
\label{table:ours_ave}
\end{table*}

\begin{table*}[!t]
% \vspace{2mm} %
\centering
\footnotesize
\setlength{\tabcolsep}{10pt}
% \resizebox{0.8\textwidth}{!}{
\begin{tabular}{cccccc}
\toprule
\textbf{Density of Control Signal} & \textbf{Joint} & \textbf{FID $\downarrow$} & \textbf{R-precision $\uparrow$ (Top-3)} & \textbf{Diversity $\rightarrow$ 9.503} & \textbf{MM-Dist $\downarrow$} \\
\midrule
25\% & Pelvis & 0.326 & 0.694 & 9.783 & 5.035 \\
50\% & Pelvis & 0.339 & 0.690 & 9.807 & 5.043 \\
75\% & Pelvis & 0.350 & 0.686 & 9.837 & 5.051 \\
100\% & Pelvis & 0.333 & 0.682 & 9.876 & 5.072 \\
\midrule
25\% & Left Foot & 0.192 & 0.688 & 9.680 & 5.011 \\
50\% & Left Foot & 0.216 & 0.694 & 9.814 & 5.020 \\
75\% & Left Foot & 0.251 & 0.689 & 9.789 & 5.013 \\
100\% & Left Foot & 0.257 & 0.681 & 9.752 & 5.028 \\
\midrule
25\% & Right Foot & 0.178 & 0.682 & 9.644 & 5.027 \\
50\% & Right Foot & 0.183 & 0.686 & 9.730 & 5.026 \\
75\% & Right Foot & 0.241 & 0.690 & 9.785 & 5.021 \\
100\% & Right Foot & 0.212 & 0.686 & 9.773 & 5.026 \\
\midrule
25\% & Head & 0.275 & 0.671 & 9.660 & 5.052 \\
50\% & Head & 0.312 & 0.669 & 9.752 & 5.171 \\
75\% & Head & 0.318 & 0.651 & 9.746 & 5.050 \\
100\% & Head & 0.288 & 0.669 & 9.766 & 5.063 \\
\midrule
25\% & Left Wrist & 0.184 & 0.676 & 9.719 & 5.013 \\
50\% & Left Wrist & 0.204 & 0.685 & 9.690 & 5.019 \\
75\% & Left Wrist & 0.214 & 0.662 & 9.711 & 5.032 \\
100\% & Left Wrist & 0.236 & 0.682 & 9.637 & 5.033 \\
\midrule
25\% & Right Wrist & 0.249 & 0.677 & 9.699 & 5.127 \\
50\% & Right Wrist & 0.252 & 0.669 & 9.658 & 5.128 \\
75\% & Right Wrist & 0.253 & 0.682 & 9.676 & 5.130 \\
100\% & Right Wrist & 0.259 & 0.671 & 9.647 & 5.115 \\
% \midrule
% Avg. & - & 0.255 & 	0.680 & 	9.735 & 	5.054 \\

\bottomrule
\end{tabular}
% }
% \vspace{-2mm} %
\caption{
The detailed re-evaluation results of OmniControl on HumanML3D.
}
\label{table:omni_ave}
\end{table*}

The visualization results in Fig.~\ref{fig:userstudy_vis} highlight several key differences between the two models.
When using only coarse-grained text, both CoMo and FineXtrol generally produce motions that align well with the input descriptions. 
However, in scenarios involving more complex or dynamic actions (\textit{e.g.}, \textit{A person is working on their boxing form.}), FineXtrol exhibits a stronger ability to generate vivid and expressive movements.
More importantly, when fine-grained textual control signals are introduced within specific temporal intervals, CoMo's adherence to the original coarse-grained text often deteriorates.
For example, given the coarse-grained text ``A person carefully kneels down, first with the right leg, then with the left one,'' CoMo mistakenly initiates the movement with the left leg.
In general, CoMo struggles with tasks that demand precise control over body parts in designated time intervals, frequently producing motions that are inconsistent with the control instructions.
In contrast, FineXtrol reliably executes the specified fine-grained control signals within the correct temporal windows.
This observation is further supported by the statistical results from our user study (see Fig.~\ref{fig:user_study} in the main paper), which confirm that FineXtrol provides significantly better fine-grained control over body part movements.

% B.3 ends here
% ===================================================================
% B.4 
\subsection{Detailed Results for Quantitative Evaluation}
\label{appendix: detailed results}
% ===================================================================
We compared two spatially controllable motion generation methods, OmniControl and InterControl, in Tab.~\ref{table:main}.(4) of the main paper, as well as the text-controllable CoMo method in Tab.~\ref{table:main}.(5), which shares the same control signal modality as our approach.
Here, we present the detailed results of a representative spatial control method, OmniControl, along with those of our FineXtrol, to examine the performance trends across different body parts under varying control signal densities.
For detailed results of CoMo, please refer to Sec.~\ref{sec:COMO_Quantitative_SM}.
To ensure fair comparisons, we re-evaluate OmniControl\cite{OmniControl_ICLR2024} using the same control signal density settings as those employed in our FineXtrol framework, enabling a consistent evaluation protocol across both methods.
In this re-evaluation, we retain the use of single-joint control signals, consistent with OmniControl’s original setup. 
It is important to note that using multiple-joint controls within a single body part can lead to degraded performance, as indicated by the trend observed in Tab.~\ref{table:omni_cross_mean}.

Interestingly, increasing the control signal density does not always benefit OmniControl—a trend also observed in its original paper.
For example, when controlling the `Pelvis' joint, a 100\% control density leads to worse performance (FID: 0.3333; Top-3 R-Precision: 0.6816) than using only 25\% density (FID: 0.3255; Top-3 R-Precision: 0.6943). 
A similar trend is observed for the `Left Foot' and `Right Foot' joints, where sparser control yields better FID scores. 
We attribute this to the challenge of simultaneously satisfying both the coarse-grained text and densely specified global joint coordinate sequences, which impose rigid constraints on the model.
Such strict control can hinder the generation of natural motions and weaken alignment with coarse-grained texts.

In contrast, our FineXtrol demonstrates more consistent improvements as control signal density increases.
It exhibits a stable upward trend in performance and consistently outperforms OmniControl across most metrics and body parts.
We believe this advantage stems from the nature of our fine-grained textual control signals, which are expressed in relative terms and thus allow more flexibility during generation.

Beyond single body part control, we also report the average performance of controlling multiple body parts across all density levels in Tab.~\ref{table:main} of the main paper.
Here, we further provide detailed multi-part control results for FineXtrol and OmniControl under the four control densities in Tab.~\ref{table:ours_cross_mean} and Tab.~\ref{table:omni_cross_mean}, respectively.
The results echo our earlier observations: as the number of control signals increases, FineXtrol generally improves, while OmniControl tends to degrade.

\begin{table}[h]
\centering
\footnotesize
\setlength{\tabcolsep}{5pt} % 适当调整列间距，默认为 6pt
\begin{tabular}{ccccc} % 删除了竖线 |
\toprule
% 使用 \thead 命令处理表头，允许用 \\ 换行
\textbf{\thead{Density of \\ Control}}  & \textbf{FID $\downarrow$} & \textbf{\thead{R-precision $\uparrow$ \\ (Top-3)}} & \textbf{\thead{Diversity $\rightarrow$ \\9.503}} & \textbf{\thead{MM-Dist $\downarrow$}} \\
\midrule
25\% &  0.271 & 0.681 & 9.681 & 5.101 \\
50\% &  0.358 & 0.677 & 9.742 & 5.127 \\
75\% &  0.428 & 0.668 & 9.648 & 5.177 \\
100\% & 0.348 & 0.679 & 9.560 & 5.178 \\

\bottomrule
\end{tabular}
\caption{
    The detailed results of our FineXtrol for controlling multiple body parts  (\textit{Cross}) on HumanML3D.
}
\label{table:ours_cross_mean}
\end{table}

\begin{table}[h]
\centering
\footnotesize % Using footnotesize instead of tiny for better readability
\setlength{\tabcolsep}{5pt} % Adjusting column separation for a better fit
\begin{tabular}{ccccc} % Removed vertical lines for booktabs style
\toprule
% Using \thead for multi-line headers
\textbf{\thead{Density of \\ Control}}  & \textbf{FID $\downarrow$} & \textbf{\thead{R-precision $\uparrow$ \\ (Top-3)}} & \textbf{\thead{Diversity $\rightarrow$ \\9.503}} & \textbf{\thead{MM-Dist $\downarrow$}} \\
\midrule
25\% &  0.620 & 0.600 & 9.297 & 5.244 \\
50\% & 0.627 & 0.604 & 9.340 & 5.258 \\
75\% &  0.616 & 0.603 & 9.357 & 5.249 \\
100\% &  0.634 & 0.595 & 9.342 & 5.256 \\
\bottomrule
\end{tabular}
\caption{
    The detailed re-evaluation results of OmniControl for controlling multiple body parts  (\textit{Cross}) on HumanML3D.
}
\label{table:omni_cross_mean}
\end{table}

% \begin{table}[!h]
% \centering
% \footnotesize
% \setlength{\tabcolsep}{1.8mm} % 减少列间距
% % \resizebox{\columnwidth}{!}{
% \begin{tabular}{c|cccc}
% \toprule
% \multirow{2}{*}{\textbf{Text Encoder}} & \multirow{2}{*}{\textbf{FID $\downarrow$}} & \multirow{2}{*}{\textbf{R-Top3}} & \textbf{Diversity $\rightarrow$} & \textbf{Matching} \\
% &  &  & \textbf{9.503} & \textbf{Score} $\downarrow$ \\

% \midrule
% % ~\cite{CLIP_ICML2021}
%  % ~\cite{T5_JMLR2020}
% Sentence-level  & 0.499 & 0.565 & 9.919 & 5.665 \\
% Snippet-level   & 0.409 & 0.570 & 9.858 & 5.645 \\

% \bottomrule

% \end{tabular}
% % }
% % \vspace{-0.5em} %
% \caption{
%     Ablation study on the impact of hierarchical contrastive training.
%     The controllable motion generation performance achieved by our text encoder, hierarchically trained across three levels, significantly outperforms the others, demonstrating the effectiveness of our hierarchical contrastive learning module in capturing fine-grained textual control signals.
%     }
% % \vspace{-1em}
% \label{table: Fine-grained textual encoder}
% \end{table}

% B.4 ends here
% ===================================================================
% B.5 

\begin{table*}[!t]
% \vspace{-6mm} %
\centering
\small 
\setlength{\tabcolsep}{12pt}
\begin{tabular}{c>{\centering}p{12mm}>{\centering}p{12mm}>{\centering}p{12mm}cccccccc}
\midrule
\multirow{2.5}{*}{\textbf{No.}} & \multicolumn{3}{c}{\textbf{Level of Contrastive Learning}} & \multicolumn{2}{c}{\textbf{Sentence level}} & \multicolumn{2}{c}{\textbf{Snippet level}} & \multicolumn{2}{c}{\textbf{Sequence level}} \\
\cmidrule(lr){2-4} \cmidrule(lr){5-6} \cmidrule(lr){7-8} \cmidrule(lr){9-10}
 & \textbf{Sentence} & \textbf{Snippet} & \textbf{Sequence} & \textbf{Pos.$\downarrow$} & \textbf{Neg.$\uparrow$} & \textbf{Pos.$\downarrow$} & \textbf{Neg.$\uparrow$} & \textbf{Pos.$\downarrow$} & \textbf{Neg.$\uparrow$} \\

\midrule
(1) & - & - & - & 0.118 & 0.276 & 0.100 & 0.228 & 0.025 & 0.235 \\
(2) & \checkmark & - & - & 0.110 & 0.596 & 0.069 & 0.446 & 0.019 & 0.334 \\
(3) & - & \checkmark & - & 0.093 & 0.589 & 0.034 & 0.590 & 0.012 & 0.404 \\
(4) & - & - & \checkmark & 0.094 & 0.603 & 0.045 & 0.542 & 0.009 & 0.407 \\
(5) & \checkmark & \checkmark & - & 0.093 & 0.594 & 0.033 & 0.594 & 0.013 & 0.384 \\
(6) & \checkmark & \checkmark & \checkmark & \textbf{0.092} & \textbf{0.625} & \textbf{0.032} & \textbf{0.611} & \textbf{0.008} & \textbf{0.422} \\

\midrule
\end{tabular}
% }
% \vspace{-2mm} %
\caption{
Ablation study on hierarchical training in the proposed contrastive learning module.
We evaluate text encoders trained with different combinations of contrastive learning levels by extracting embeddings for 10,000 randomly sampled positive (Pos.) and negative (Neg.) pairs from each level of textual descriptions, and computing their average cosine similarity.
The results indicate that incorporating all three levels of contrastive learning enables the text encoder to best capture the semantics of fine-grained textual descriptions, resulting in better embeddings of our fine-grained textual control signals.
}
\label{table:CL_cosine}
\end{table*}

\begin{table*}[!t]
\centering
\small 
% Adjust column separation to fit the two-column format
\setlength{\tabcolsep}{16pt} 
\begin{tabular}{c ccc  cccc}
\toprule
% --- HEADER ROW 1: Main Titles ---
\multirow{2.5}{*}{\textbf{No.}} & \multicolumn{3}{c}{\textbf{Level of Contrastive Learning}} & \multirow{2.5}{*}{\textbf{FID $\downarrow$}} & \multirow{2.5}{*}{\textbf{R-Top3} $\uparrow$} & \textbf{Diversity $\rightarrow$} & \multirow{2.5}{*}{\textbf{MM-Dist $\downarrow$}} \\

% --- HEADER ROW 2: Sub-Titles and rules ---
\cmidrule(lr){2-4}
& \textbf{Sentence} & \textbf{Snippet} & \textbf{Sequence} & & & \textbf{9.503} &  \\

\midrule
% --- DATA ROWS ---
% The checkmarks below correspond to rows (1), (2), and (3) from your second source table.
% You can adjust them to match what each data row represents.
(i) & \checkmark  & -          & -          & 0.499          & 0.565          & 9.919          & 5.665 \\
(ii) & \checkmark & \checkmark          & -          & 0.409          & 0.570          & 9.858          & 5.645 \\
(iii) & \checkmark          & \checkmark & \checkmark          & \textbf{0.245} & \textbf{0.685} & \textbf{9.492} & \textbf{5.087} \\

\bottomrule
\end{tabular}
\caption{
    Ablation study on the impact of hierarchical contrastive training.
    The controllable motion generation performance achieved by our text encoder, hierarchically trained across three levels, significantly outperforms the others, demonstrating the effectiveness of our hierarchical contrastive learning module in capturing fine-grained textual control signals.
    }
\label{table: Fine-grained textual encoder} % Using a new label for the combined table
\end{table*}

\subsection{More Ablation Study}
\label{appendix: ablation study}
% ===================================================================
% \vspace{0.5em}
\subsubsection{Effectiveness of the Hierarchical Training in the Proposed Contrastive Learning Module.}
Fig.~\ref{fig:text_encoder_comparison} in the main paper compares our text encoder—trained with three-level hierarchical contrastive learning—with two widely used encoders, CLIP and T5.
In this subsection, we further analyze the impact of each level on the quality of fine-grained textual embeddings, as shown in Tab.\ref{table:CL_cosine}.
Specifically, we use text encoders trained with different combinations of contrastive learning levels to extract embeddings for 10,000 randomly sampled positive and negative pairs from each level of textual descriptions. 
We then compute the average cosine similarity for each group.
A more effective encoder for our fine-grained text controllable motion generation task should yield lower cosine similarity for negative pairs and higher similarity for positive pairs, indicating better semantic discrimination.
The results demonstrate that applying contrastive learning at a single level (Rows 2–4) enhances the embeddings at the sentence, snippet, and sequence levels compared to the original T5 encoder (Row 1).
Further improvements are observed when multiple contrastive learning levels are combined (Rows 5 and 6).
Our final encoder, trained with all three levels of hierarchical contrastive learning (Row 6), achieves the best overall performance in capturing fine-grained textual semantics and is therefore adopted as the default in all experiments.

\subsubsection{Impact of Hierarchical Contrastive Training on Fine-Grained Text Controllable Motion Generation.}
We further investigate how different levels of contrastive learning applied to the text encoder affect the controllability of generated motions.
Specifically, we evaluate the performance of our fine-grained text-controllable motion generation task using text encoders trained with (i) sentence-level contrastive learning, and (ii) both sentence- and snippet-level contrastive learning, as shown in Tab.~\ref{table: Fine-grained textual encoder}.
The results demonstrate that the encoder trained with (ii) consistently outperforms the one trained with (i) (e.g., FID: 0.409 vs. 0.499). 
However, both configurations yield significantly lower performance compared to the encoder trained with all three levels, underscoring their limited ability to capture fine-grained textual control signals.
These findings highlight the importance of hierarchical contrastive training across all three levels for accurately interpreting fine-grained text and generating high-quality motions.

\subsubsection{Optimal Temperature and Pooling Methods.} 

Tab.\ref{table:token_temp_CL} investigates the impact of different pooling strategies for textual embeddings $\text{Avg}(\cdot)$ in Equation~\ref{eqn:text_encoder}, as well as the effect of the temperature parameter $\tau$ in Equation~\ref{eqn:infoNCE}.
Specifically, we use a text encoder trained with sentence-level contrastive learning to extract embeddings for 10,000 randomly sampled sentence-level positive pairs, and compute their mean cosine distance.
The results show that averaging all output tokens from the text encoder yields better performance than using only the first output token as the sentence embedding.
Additionally, the best performance—indicated by the smallest cosine distance between positive pairs—is achieved when the temperature is set to $\tau = 0.07$.

\begin{table}[!h]
\centering
% \vspace{-2mm}
\small
\begin{tabularx}{\columnwidth}{p{2cm}XXXXX}
\toprule
\multirow{2.5}{*}{\makecell{\textbf{Textual} \\ \textbf{Embedding}}} & \multicolumn{5}{c}{\textbf{Temperature $\tau$}} \\
\cmidrule(lr){2-6}
 & \textbf{\centering 0.01} & \textbf{0.03} & \textbf{0.05} & \textbf{0.07} & \textbf{0.10} \\
\midrule
Average token & 0.125 & 0.121 & 0.112 & \textbf{0.110} & 0.134 \\
First token & 0.208 & 0.164 & 0.150 & 0.148 & 0.186 \\

\bottomrule
\end{tabularx}
\caption{
    Ablation study on temperature $\tau$ and pooling methods in the proposed contrastive learning module. 
    The results are the average cosine distance between textual embeddings of 10,000 randomly sampled sentence-level positive pairs, where lower values indicate better performance.
    The results suggest that average pooling yields more effective textual embeddings, and the optimal $\tau$ is 0.07.
}
\label{table:token_temp_CL}
\end{table}

\begin{table}[!h]
\centering
\footnotesize
\setlength{\tabcolsep}{1.6mm} % 减少列间距
% \resizebox{\columnwidth}{!}{
\begin{tabular}{c|cccc}
\toprule
\multirow{2}{*}{\textbf{Masking Prob.}} & \multirow{2}{*}{\textbf{FID $\downarrow$}} & \multirow{2}{*}{\textbf{R-Top3}} & \textbf{Diversity $\rightarrow$} & \multirow{2}{*}{\textbf{MM-Dist $\downarrow$}} \\
&  &  & \textbf{9.503} & \\

\midrule
% ~\cite{CLIP_ICML2021}
 % ~\cite{T5_JMLR2020}
Varying  & 0.408 & 0.560 & 9.944 & 5.614 \\

Fixed 0.5 & \textbf{0.245} & \textbf{0.685} & \textbf{9.492} & \textbf{5.087} \\

\bottomrule

\end{tabular}
% }
% \vspace{-0.5em} %
\caption{
    Ablation study on the masking probability in fine-grained textual control signals during training.
    }
% \vspace{-1em}
\label{table:final_version}
\end{table}

\subsubsection{Masking Probability in Fine-Grained Textual Control Signals.} 

As described in Sec.~\ref{appendix: implementation details}, we randomly replace descriptions within temporal intervals with the special token \texttt{<Mask>} at a fixed probability of 50\% during training.
Here, we further investigate the effect of randomly masking these descriptions with \textit{varying} probabilities.
As shown in Tab.~\ref{table:final_version}, the \textit{varying} probabilities results in significantly worse motion quality and text-motion alignment.
While it yields a slightly higher diversity score, the reduced alignment indicates a lack of control over the generated motions.
These results highlight the effectiveness of our fixed 50\% masking strategy.

\begin{table*}[!t]
\begin{center}
\footnotesize
\setlength{\tabcolsep}{8pt}
\begin{tabular*}{0.95\textwidth}{@{\extracolsep{\fill}}cc c c cc}
    \toprule[1pt]
    \textbf{Density of Control Signal} &
    \textbf{Body Part}  &
    \textbf{FID $\downarrow$} &
    \textbf{R-precision $\uparrow$ (Top-3)} &
    \textbf{Diversity $\rightarrow$ 9.503} &
    \textbf{MM-Dist $\downarrow$} \\
    \midrule[0.5pt]
    - & - & 0.544  & 0.611 & 9.559 & 5.432 \\
    \midrule[0.5pt]

    25\% & \multirow{4}{*}{Head} & 1.339 & 0.581 & 9.337 & 4.199 \\
    50\% &  & 1.391 & 0.562 & 9.259 & 4.397 \\
    75\% &  & 1.033 & 0.613 & 9.045 & 4.147 \\
    100\%  &  & 1.251 & 0.593 & 9.225 & 4.256 \\
    \midrule[0.5pt]
    25\% & \multirow{4}{*}{Body} & 1.044 & 0.635 & 9.360 & 3.929 \\
    50\% &  & 1.096 & 0.595 & 9.456 & 4.207 \\
    75\% &  & 0.969 & 0.619 & 9.054 & 4.072 \\
    100\%  &  & 1.002 & 0.645 & 9.133 & 4.151 \\
    \midrule[0.5pt]
    25\% & \multirow{4}{*}{Left Hand} & 1.356 & 0.609 & 9.241 & 3.967 \\
    50\% &  & 1.295 & 0.606 & 9.160 & 4.090 \\
    75\% &  & 1.197 & 0.619 & 9.060 & 4.027 \\
    100\%  &  & 1.528 & 0.610 & 9.265 & 4.136 \\
    \midrule[0.5pt]
    25\% & \multirow{4}{*}{Right Hand} & 1.920 & 0.608 & 9.495 & 4.258 \\
    50\% &  & 1.968 & 0.556 & 9.514 & 4.421 \\
    75\% &  & 1.382 & 0.586 & 8.941 & 4.275 \\
    100\%  &  & 1.484 & 0.605 & 9.223 & 4.197 \\
    \midrule[0.5pt]
    25\% & \multirow{4}{*}{Left Leg} & 1.216 & 0.628 & 9.420 & 3.903 \\
    50\% &  & 1.340 & 0.611 & 9.151 & 4.123 \\
    75\% &  & 1.301 & 0.612 & 9.014 & 4.080 \\
    100\%  &  & 0.940 & 0.623 & 9.198 & 4.087 \\
    \midrule[0.5pt]
    25\% & \multirow{4}{*}{Right Leg} & 2.038 & 0.582 & 9.506 & 4.337 \\
    50\% &  & 2.169 & 0.558 & 9.269 & 4.466 \\
    75\% &  & 1.449 & 0.589 & 9.239 & 4.298 \\
    100\%  &  & 1.486 & 0.581 & 9.225 & 4.269 \\
    
    \bottomrule[1pt]
    
\end{tabular*}
\caption{
    The detailed results of `Direct' Paradigm on HumanML3D.
} 
\label{table:MDM_fine_grained_SM}
\end{center}
\end{table*}

\subsubsection{Detailed Results of `Direct' Paradigm.} 
\begin{figure}[!h]
\begin{center}
\includegraphics[width=1.0\linewidth]{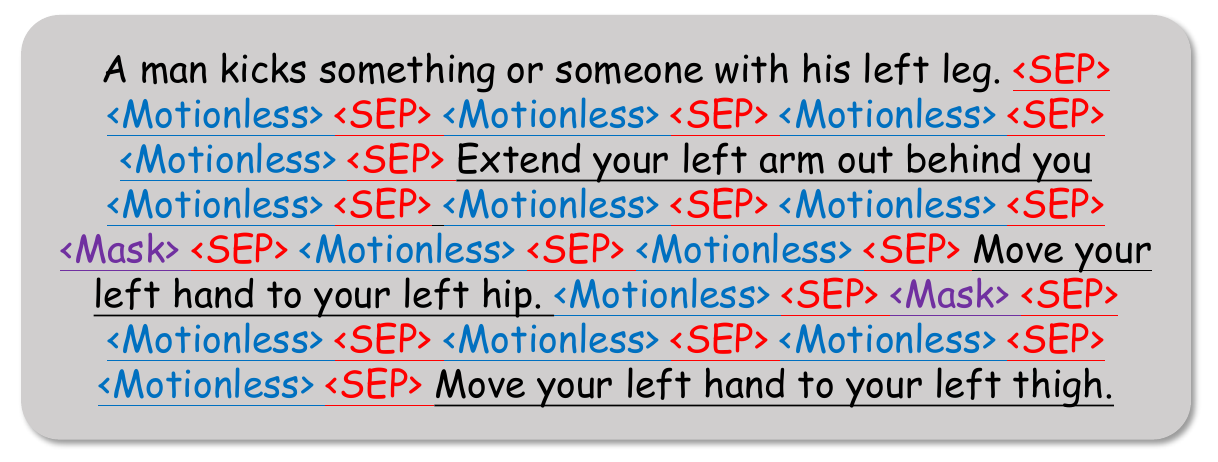}
\end{center}
\setlength{\abovecaptionskip}{-0em}  % Space above the caption
\caption{
    An example input of the `Direct' paradigm.
    The first sentence is the coarse-grained text, while the underlined sentences indicate the fine-grained textual control signal.
}
\label{fig:mdm_dt_example}
\end{figure}

In Sec.~\ref{sec:ablation_study}, we investigate replacing our control paradigm with the `Direct' paradigm, which merges the fine-grained textual control signals with the coarse-grained text into a single input. 
An example of such input is shown in Fig.~\ref{fig:mdm_dt_example}. 
The average performance of controlling different body parts under four levels of control signal density is reported in Tab.~\ref{table:Ablation_FG_MDM} in the main paper.
Here, we present the detailed results for each body part under four control density levels in Tab.~\ref{table:MDM_fine_grained_SM}.
As the control signal density increases, the model’s performance on motion generation metrics gradually improves. 
However, both the FID and MM-Dist degrade notably compared to our FineXtrol method, even performing worse than the original MDM baseline without control signals (Row 1).

% A crucial element of our  FineXtrol is the dual-branch ControlNet architecture, which processes coarse-grained texts and fine-grained textual control signals separately. To validate our dual-branch architecture, we experimented with a single-branch baseline. This baseline adapts the original MDM framework by encoding both the coarse-grained prompt and the fine-grained descriptions together through a single, unified CLIP encoder. An example of this concatenated input format is shown in Figure~\ref{fig:mdm_dt_example}.

\subsection{More Visualization Results}
\label{appendix: Visualization}

\begin{figure*}
\begin{center}
\includegraphics[width=1.0\textwidth]{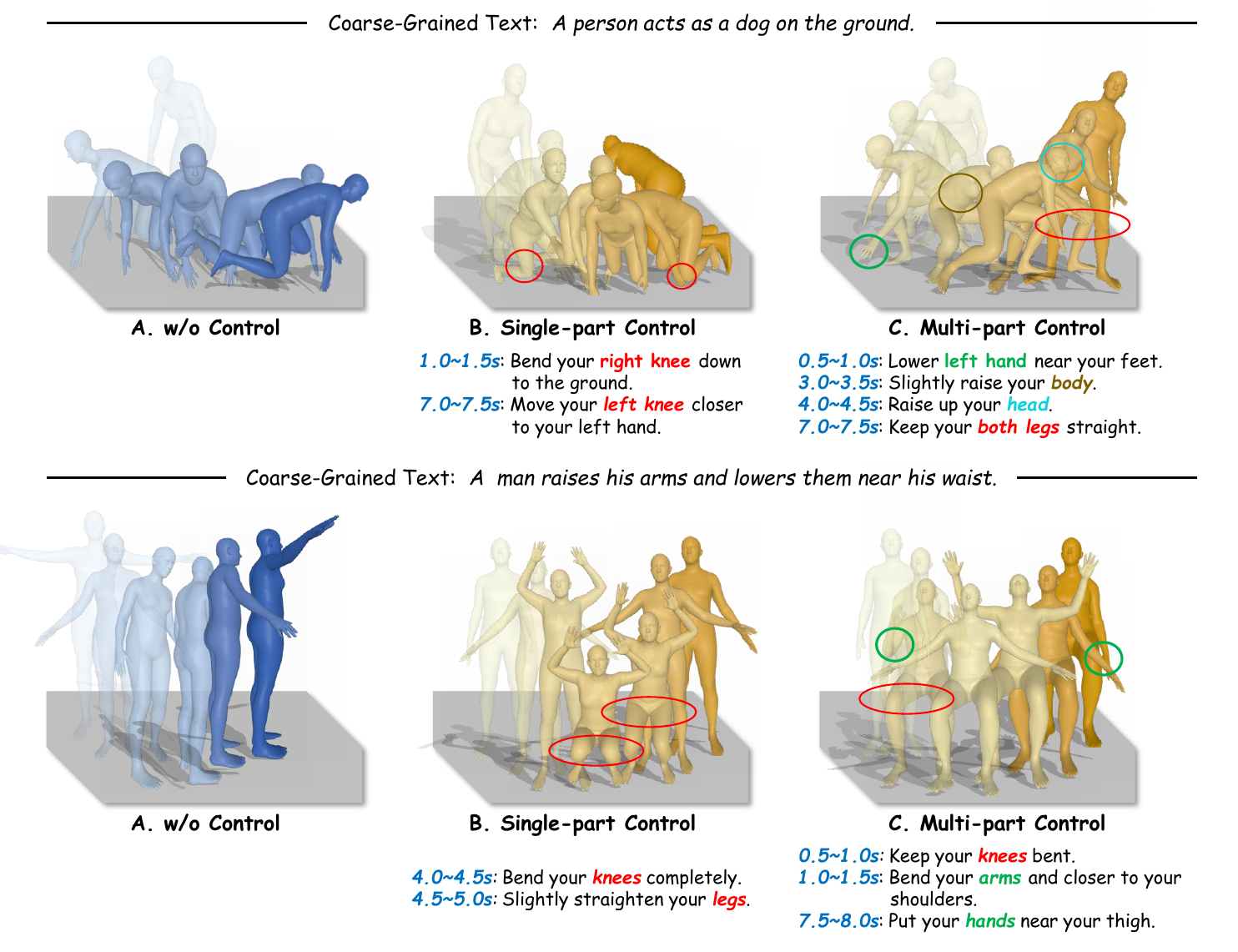}
\end{center}
\caption{
    More visualizations of different control settings on the HumanML3D~\cite{T2M_CVPR2022} test set. 
    \texttt{<Mask>} is used for all unspecified temporal intervals. 
    FineXtrol successfully generates realistic human motions conditioned on various fine-grained textual control signals.
    Within each sequence, darker colors represent later frames.
}
\label{fig:append_more_visual}
\end{figure*}

Fig.\ref{fig:append_more_visual} presents additional visualizations of motions generated by our FineXtrol on the HumanML3D~\cite{T2M_CVPR2022} test set. 
From left to right, the columns show generated motions without control signals, with control over a single body part, and with controls over multiple body parts—all conditioned on the same coarse-grained texts.

% B.6 ends here
% ===================================================================
% B.7 
\subsection{Details of User Study}
\label{appendix: User Study}

\begin{figure*}[!t]
\begin{center}
\includegraphics[width=0.8\textwidth]{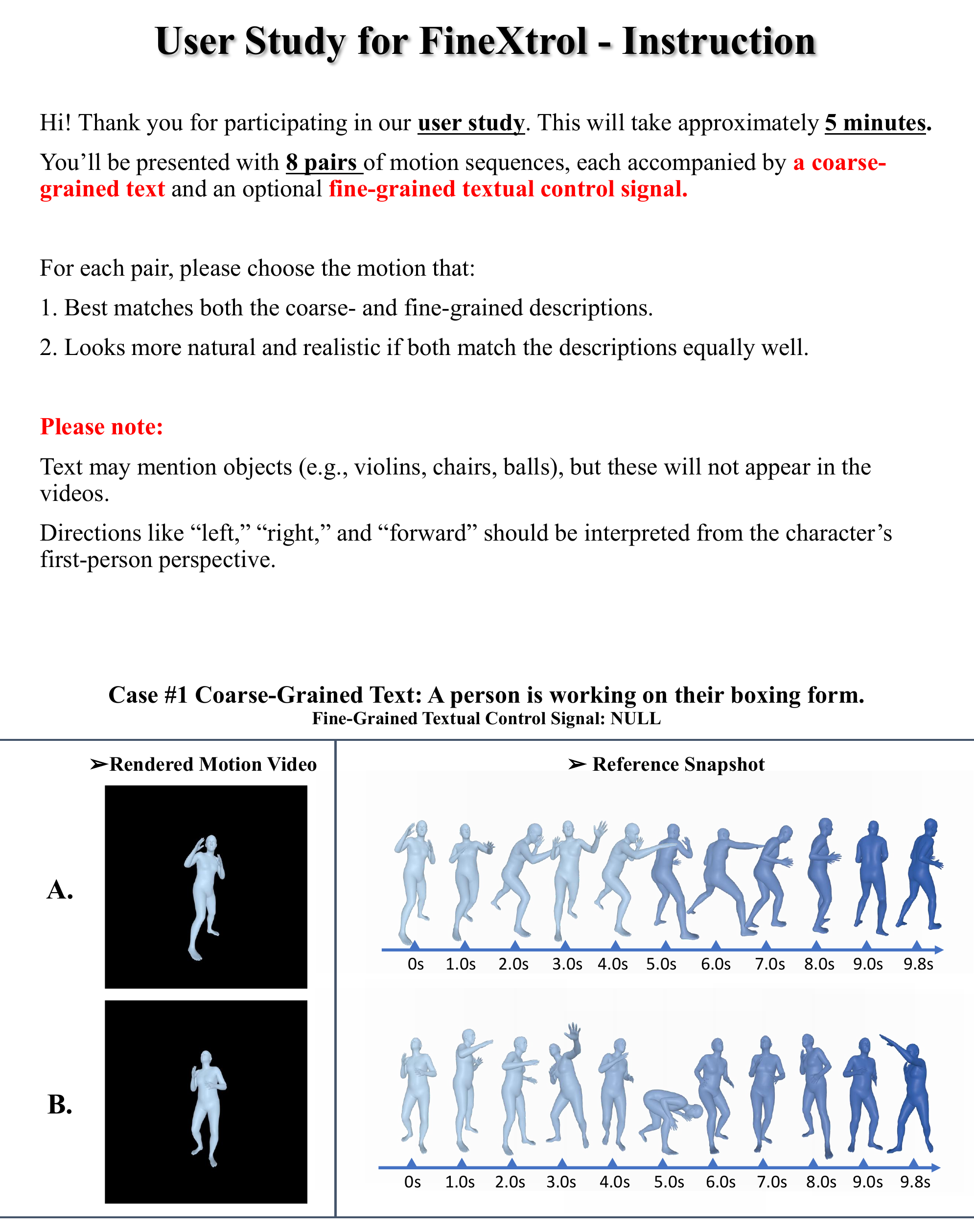}
\end{center}
\caption{
    The evaluation instructions and one of the video pairs presented to participants in our user study.}
\label{fig:append_userstudy_ins}
\end{figure*}

In our user study, we conducted a perceptual evaluation to compare FineXtrol with CoMo~\cite{CoMo_ECCV2024}.
We generated 8 pairs of 3D motion videos (16 in total) using coarse-grained texts from the HumanML3D~\cite{T2M_CVPR2022} test set.
These samples cover different fine-grained textual control settings, including 2 examples without control signals and 6 with control signals over different body parts. 
Participants were shown each video pair and asked to choose the one that better matched both the coarse-grained text and the fine-grained control signals.
Fig.~\ref{fig:append_userstudy_ins} illustrates the evaluation instructions and one of the video pairs presented to participants.
Full user study are available in the PowerPoint file included in the supplementary materials.

\FloatBarrier
% \clearpage 

% B.5 ends here
% ===================================================================
% B.6 
\onecolumn

% B.7 ends here
% ===================================================================
% Appendix ends 

\twocolumn

\end{document}